\documentclass[default,iicol]{sn-jnl}

\jyear{2021}%

\usepackage{float}
\usepackage{graphicx}
\usepackage{hyperref}       %
\hypersetup{
    colorlinks,
    linkcolor={blue},
    citecolor={blue},
    urlcolor={blue}
}
\usepackage{amsfonts,dsfont,amsmath}
\usepackage{multirow}
\usepackage{multicol}
\usepackage{totcount} 
\newtotcounter{citnum} 
\def\oldbibitem{} \let\oldbibitem=\bibitem
\def\bibitem{\stepcounter{citnum}\oldbibitem}
\newcommand{\bmd}{{\boldsymbol d}}

\newcommand{\bmX}{{\boldsymbol X}}
\newcommand{\bmx}{{\boldsymbol x}}
\newcommand{\bmY}{{\boldsymbol Y}}

\theoremstyle{thmstyleone}%
\theoremstyle{thmstyletwo}%

\theoremstyle{thmstylethree}%

\begin{document}

\title[FAD]{Functional Anomaly Detection: a Benchmark Study}


\author*[1]{\fnm{Guillaume} \sur{Staerman}}\email{guillaume.staerman@telecom-paris.fr}
\author[1]{\fnm{Eric} \sur{Adjakossa}}
\author[1]{\fnm{Pavlo} \sur{Mozharovskyi}}

\author[2]{\fnm{Vera} \sur{Hofer}}

\author[3]{\fnm{Jayant} \sur{Sen Gupta}}
\author[1]{\fnm{Stéphan} \sur{Clémençon}}

\affil*[1]{\orgdiv{LTCI}, \orgname{Télécom Paris, Institut Polytechnique de Paris}, \orgaddress{ \country{France}}}

\affil[2]{\orgdiv{Department of Operations and Information Systems}, \orgname{University of Graz}, \orgaddress{ \country{Austria}}}

\affil[3]{ \orgname{Airbus AI Research}, \orgaddress{ \city{Toulouse}, \country{France}}}



\date{Received: date / Accepted: date}

\abstract{The increasing automation in many areas of the Industry expressly demands to design efficient machine-learning solutions for the detection of abnormal events. With the ubiquitous deployment of sensors monitoring nearly continuously the health of complex infrastructures, anomaly detection can now rely on measurements sampled at a very high frequency, providing a very rich representation of the phenomenon under surveillance. In order to exploit fully the information thus collected, the observations cannot be treated as multivariate data anymore and a functional analysis approach is required. 
It is the purpose of this paper to investigate the performance of recent techniques for anomaly detection in the functional setup on  real datasets. After an overview of the state-of-the-art and a visual-descriptive study, a variety of anomaly detection methods are compared. While taxonomies of abnormalities (\textit{e.g.} shape, location) in the functional setup are documented in the literature, assigning a specific type  to the identified anomalies appears to be a challenging task. Thus, strengths and weaknesses of the existing approaches are benchmarked in view of these highlighted types in a simulation study. Anomaly detection methods are next evaluated on two datasets, related to the monitoring of helicopters in flight and to the spectrometry of construction materials namely. The benchmark analysis is concluded by a recommendation guidance for practitioners.}

\keywords{Anomaly detection, Functional data,  Depth statistics,  Taxonomy of anomalies,  Benchmark study}



\maketitle

\section{Introduction}\label{intro}

The infatuation with machine learning is spreading to nearly all fields (\textit{e.g.} science, transportation,
energy, medicine, security, banking, insurance, commerce) as the ubiquity of sensors (\textit{e.g.} IoT) make more and more data at disposal with an ever finer granularity. An abundance of new applications such as the health monitoring of complex infrastructures (\textit{e.g.} aircrafts, energy networks) together with the availability of
massive data samples (Big Data) and technological bricks for information acquisition and
access (\textit{e.g.} sensor networks, IoT, distributed file systems) and  computation (\textit{e.g.} infrastructures for
massively parallelized/distributed computation, on-line processing) has put pressure on the scientific community to develop new artificial intelligence methods and algorithms, following the example of successful applications such as computer vision, machine-listening or machine language translation. In particular, (unsupervised) anomaly  detection remains a very active research topic. It can be stated in an informal manner as the identification of data points $x$ that are suspicious due to their significant difference from the vast majority of observed data. The goal pursued is the same as in binary classification, to build a classification rule $g$ mapping the input space, $\mathcal{X}$ say, to $\{-1,+1 \}$ namely: an observation $x\in \mathcal{X}$ is then considered as abnormal/anomalous, \textit{i.e.} not distributed according to the same probability measure $P$ as that of the 'normal' (to be understood as 'not abnormal', with no reference to the Gaussian distribution here) observations when $g(x)=+1$, and as normal when $g(x)=-1$. However, in contrast to the supervised situation where a binary label is assigned to each training observation indicating whether it is normal, only unlabeled data, supposedly normal for the most part, are available for learning the decision rule $g$ in the unsupervised case.
With the deployment of sensors monitoring their operation nearly continuously, machine-learning is expected to provide solutions for
predictive maintenance of sophisticated systems, such as electricity grids or aircrafts, facilitating
the early detection of “weak” signals that announce breakdowns, and serving to plan the replacement of components before their probable failure. Beyond the control of the false alarm rate, the challenge now essentially consists in fully exploiting the information collected, taking often the form of measurements of a physical variable sampled at a very high frequency. In this case, the observations cannot be treated as multivariate data and a functional approach is required. The (unsupervised) anomaly detection issue in the multivariate setting (when $\mathcal{X}\subset \mathbb{R}^d$ with $d\geq 1$) is well-documented in the litterature and a large variety of dedicated techniques (way too numerous to be listed exhaustively here) have been proposed and investigated. The vast majority of the heuristics considered consist in turning supervised learning methods into unsupervised approaches, where label is replaced by rarity, anomalies being supposed to be rare by definition (\textit{e.g.} SVM becoming one-class SVM, classification trees becoming isolation trees). In contrast, there has not been as much attention paid to the functional situation until now, see \textit{e.g.} \cite{rousseeuw2018anomaly} or \cite{staerman2019functional} and the references therein. 
Functional data are available in the form of functions, images and shapes or more general objects \cite{wang2016functional}, their statistical analysis, referred to as functional data analysis (FDA in abbreviated form) has received much interest in the last two decades, see \textit{e.g.} \cite{ramsay2004functional}, 
\cite{ferraty2006nonparametric}, \cite{wang2016functional}.
In its simplest form, a functional dataset is a collection of curves observed on a grid $t_1<\dots<t_p$, generally supposed to possess some smoothness properties to allow their recovery by means of appropriate interpolation/approximation schemes with a controlled error, see \textit{e.g.} \cite{ramsay2007applied} for an account of FDA techniques. Though a very rich information can be carried by functional data, the downsides of FDA are challenges for designing feasible numerical procedures and for establishing a sound validity theoretical framework alike. Of course, anomaly detection for functional data can be reduced to the implementation of its counterpart in the multivariate case by means of straightforward dimensionality reduction techniques. Indeed, one may first project the observed curves onto a subspace of finite dimension by keeping the most informative components of a Kahrunen-Lo\`eve decomposition (\textit{i.e.} those with largest empirical variance) or by truncating their expansion in a Hilbertian basis of reference, using a flexible dictionary of functions (\textit{e.g.} Fourier basis, wavelets). This step is often referred to as the \textit{filtering} stage. Next, any anomaly detection method designed for multivariate observations can be used for analyzing the (parsimonious) filtered data thus obtained, see, \textit{e.g.} \cite{ferraty2006nonparametric}.
Though easy to implement, such a two-step strategy is unfortunately dramatically conditioned by the finite-dimensional representation method chosen, causing the loss of patterns possibly crucial to discriminate between normal observations and abnormal observations of certain types.
Indeed, a taxonomy of functional outliers in \cite{hubert2015multivariate} distinguish between \textit{isolated outliers}, which manifest an outlying behavior during a very short time interval, like spikes or peaks in univariate curves, and \textit{persistent outliers}, which exhibit a deviation from the normal behavior on a large part of the domain and consist of \textit{shift outliers} (same shape as the majority of the observations, but moved away), \textit{shape outliers} (curves with a shape which differs from that of the majority of the observations) or \textit{amplitude outliers} (curves with a scale/range which differs from that of the majority of the observations). Hence, a certain finite-dimensional representation tailored to the detection of a specific type of anomaly, it may completely fail in detecting the other types.
Because there exists a variety of functional anomalies, robust statistical methods such as those based on depth statistics have recently received attention, see \textit{e.g.} \cite{cuevas2007robust} or \cite{staerman2019area}.
The depth of a point within a data cloud was first introduced in \cite{tukey1975mathematics}, in order to define a notion of multivariate median in $\mathbb R^d$ with $d\geq 2$. Afterwards, so as to measure the depth of an arbitrary point $ x$, \cite{donoho1992breakdown} considered hyperplanes through $x$ and determined its depth by the smallest portion of data that are separated by such a hyperplane. Since then, this idea has proved to be very fruitful and has lead to a rich statistical methodology, still in progress, in particular with the design of more general nonparametric depth statistics. In a nutshell, the depth of a data point describes its \textit{outlyingness}, or \textit{centrality} inversely, relatively to the data cloud, and thus defines a preorder on the feature space: the smaller its depth, the likelier a point can be considered as an outlier. For a dataset at hand, in the majority of cases computation of depth of a point consists in optimising a specific function: \textit{e.g.} minimization of the directional outlyingness, since an observation is considered abnormal if it is abnormal at least in one direction, while normal observations are those that are normal in all directions. Although depth functions can be computed for empirical distributions only from a practical perspective, the consistency results documented in the literature provide guarantee that statistical versions may provide accurate estimates of the depth of the true distribution of the data. Extension of these methods and results to data in functional spaces have been recently the subject of attention, see \textit{e.g.} \cite{becker2014robustness}, with applications to functional anomaly detection in particular, refer to \textit{e.g.}, \cite{hubert2015multivariate} or \cite{nagy2017depth}.

It is the goal of this article to investigate the performance of recent techniques for functional anomaly detection and compare their accuracy with that of simpler approaches, based on a preliminary dimensionality reduction, standing as natural competitors. In particular, specific attention is paid to those that are based on functional depth statistics or that extend multivariate methods by avoiding the filtering step. A benchmark study comparing the merits of the methods considered here regarding various metrics of reference is thus presented on aeronautics data gathered by Airbus and spectrometry measurements of sedimentary material collected by the Geological Survey of Austria for quality assessment on mining sites of Austria. Specifically, the aeronautics dataset consists of one-minute-sequences of accelerometer data measured at a 1024Hz frequency. Airbus dataset is divided into two parts: the training set composed of $1677$ curves with no available labels that may contains 'abnormal' observations and the validation/test set composed of 2511 time-series with 1794 'normal' curves.  In contrast, the test data are labeled, in order to evaluate the performance of the anomaly detection rules learned in the training stage. These measurements were made on test helicopters at various locations, in various angles, on different flights. The learning framework is unsupervised: in the experiment, all accelerometer data series at disposal for training automatically a classifier to detect abnormal changes are considered as normal. The spectrometry of rocks data consists of one dataset of 2096 curves with 60000 measurements. It represents materials of two types whose labels are available, with limestone being the desirable (normal) rock type and the intrusive (abnormal) cellular dolomite. The task is thus, given the reflectance spectrum (with noise subtracted and normalized with respect to the reference spectrum) of multiple samples of mined examples, separate those abnormal.

As revealed by the experimental analysis carried out, recent (depth-based) functional anomaly detection techniques significantly outperform traditional methods. Additional experiments based on simulation data also provide empirical evidence that their flexibility permits to detect functional anomalies of different types, avoiding the limitations due to the exploitation of a specific finite-dimensional representation of the data.

The paper is organized as follows. Section~\ref{sec:methods} gives a general overview of recent anomaly detection methods for functional data, and discusses both their strengths and weaknesses. In Section~\ref{sec:simulation}, the performance metrics used for measuring the accuracy of anomaly detection rules learned in an unsupervised manner on (labeled) test data are described, the experiments on synthetic/real data and the results obtained are presented at length in Section~\ref{sec:benchmark}. Finally, some concluding remarks are collected in Section~\ref{sec:concl}. 

\section{Anomaly Detection for Functional Data}\label{sec:methods}

We start with recalling briefly the rationale behind classic approaches to functional anomaly detection and next describe more recent techniques dedicated to this task, coping directly with the functional nature of the observations.

\subsection{Classic Approaches - Filtering Techniques}\label{multi}

In the unsupervised learning framework, no label indicating whether a training observation is anomalous or not is available. Hence, anomalies should be identified in an automatic way by learning the 'normal' behavior, that of the vast majority of the observations, and considering those differing significantly from it as 'abnormal'. Logically, anomalies are rare in the data and thus fall in 'low density' regions: anomaly detection thus boils down to identifying the 'tail' of the distribution.

In the standard multivariate setup, \textit{i.e.} when the observations can be represented as points in $\mathbb{R}^d$, a number of methods have been developed in the statistical literature. When the data generating process is known, a model based anomaly detection can be used of course. For example, if the distribution modelling the normal behavior is supposed to be Gaussian, observations outside a certain ellipsoid (corresponding to an isodensity curve) are considered as abnormal. In order to recover it, a robust mean and covariance estimation should be performed, see \textit{e.g.} \cite{RousseeuwD99}. However, in most practical situations, the probability distribution of the normal observations is unknown and nonparametric approaches like Minimum Volume Set estimation (MV-set)  \cite{Polonik97,ScottN06}, One-Class Support Vector Machines (OCSVM) \cite{SchoelkopfPSTSW05}, Local Outlier Factor (LOF \cite{BreunigKNS00}) or Isolation Forest (IF) \cite{LiuTZ08,HaririCKB19} are widely used. Most of these algorithms produce, under appropriate conditions, estimates of upper-level sets of the density of the normal observations, tagging as anomalies the observations lying outside these regions. The concept of statistical depth, originally introduced by \cite{tukey1975mathematics} (refer also to \cite{ZuoS00} for a list of desirable properties a depth function should fulfill), permits to extend the notion of quantile to distributions $P$ on $\mathbb{R}^d$ when $d\geq 2$ and can also be used for the purpose of anomaly detection. By defining a center-outward ordering of points in the support of a probability distribution $P$, a depth function $D(\cdot \mid P):\mathbb{R}^d\to \mathbb{R}^d$ serves to quantify the centrality of any point $x\in \mathbb{R}^d$: points with highest depth values are ideally those near the 'center' of the mass, while observations with depth values less than a certain critical threshold (\textit{i.e.} outside a \textit{depth region} $\{x\in \mathbb{R}:\; D(\cdot \mid P)\geq \tau\}$ with $\tau>0$) are considered as abnormal, see \textit{e.g.}  \cite{Mosler13}.

\begin{figure}[!h]
\begin{center}
\begin{tabular}{cc}
\quad Properly reconstructed function \\
\includegraphics[scale=0.5,trim=0.5cm 0.5cm 1cm 0.75cm,clip=true]{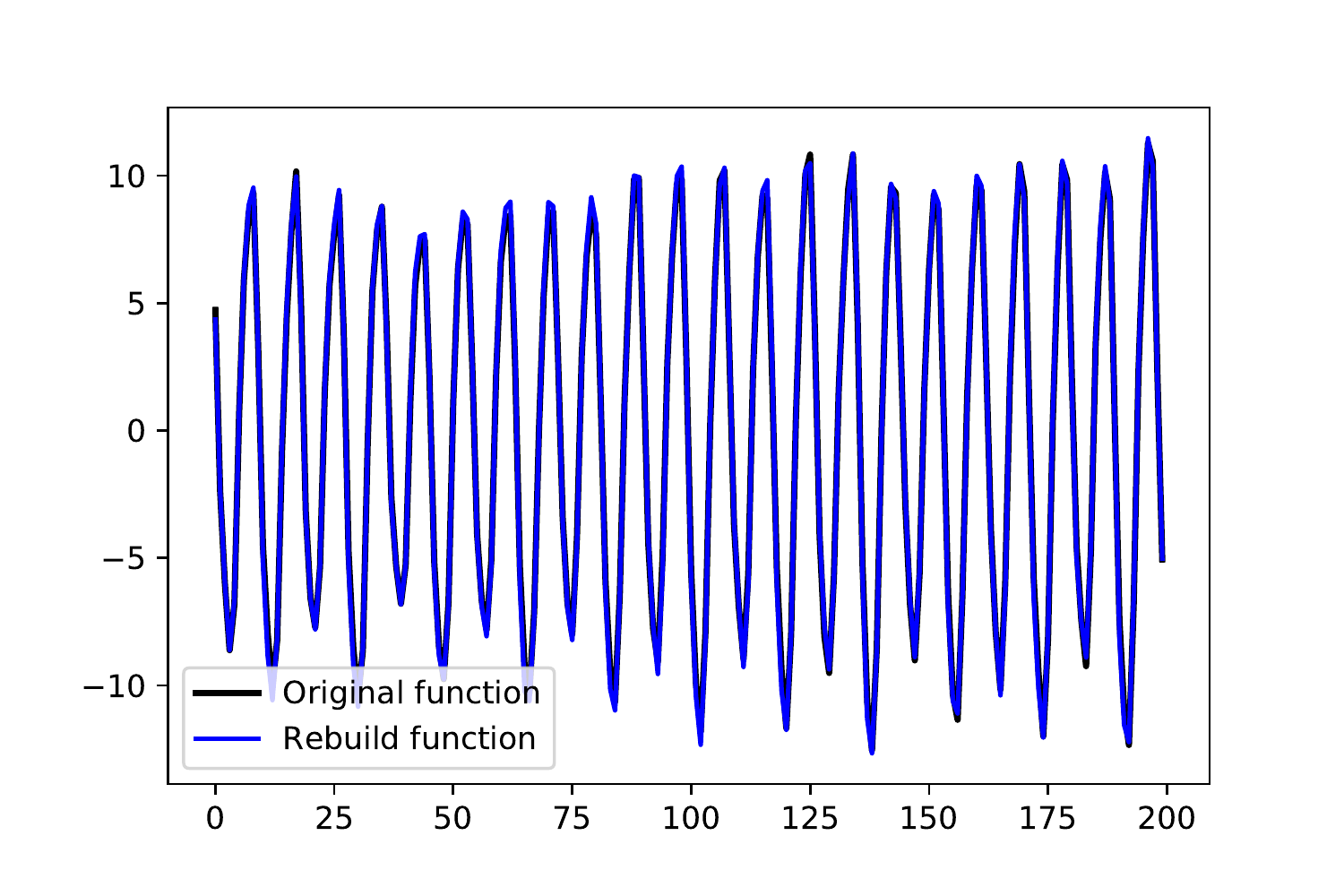}\\
Poorly reconstructed function\\
 \includegraphics[scale=0.5,trim=0.5cm 0.5cm 1cm 0.75cm,clip=true]{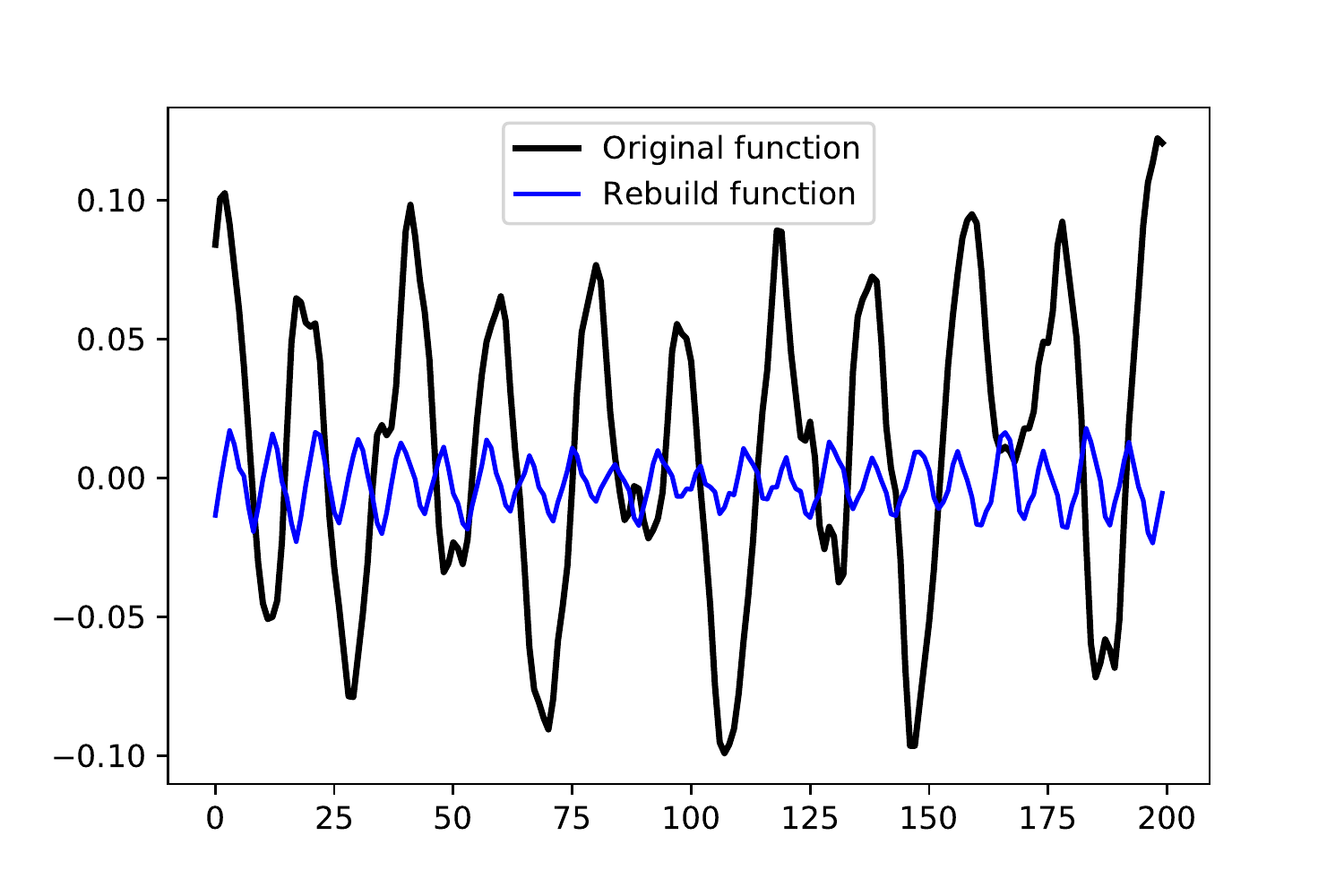}
\end{tabular}
\end{center}
\caption{Two functional observations (black curves) from the aeronautics dataset and their reconstruction (blue curves) using $10$ first principal components of FPCA: a normal observation, well reconstructed (top and an abnormal one, poorly reconstructed (bottom).}\label{fpcareconst}
\end{figure}

In order to extend the application of the techniques mentioned above to functional data, a filtering step can be used. Assuming that the variable observed takes its values in a subset of a Hilbert space, the data are first projected onto a finite dimensional subspace defined by truncating their expansion in an orthonormal basis and any multivariate anomaly detection technique can be next fed with the filtered data. The basis considered may be picked in a dictionary (\textit{e.g.} Fourier, wavelets, splines) or learned from the data as in Functional Principal Component Analysis (FPCA in abbreviated form). In the ideal case where one knows in advance which type of anomalies one attempts to detect, accuracy can be optimized by tuning the parameters of the filtering stage (basis, subspace which the data are projected onto) accordingly. However, in most situations encountered in practice, there are no guarantees about the nature of future anomalies: if the finite-dimensional representation produced by the filtering procedure can be appropriate to detect certain types of anomalies, it may completely fail in detecting the other types, see Figure~\ref{fpcareconst}. To remedy this  drawback, certain methods tailored to functional data have been recently developed.

Table~\ref{tab:benchfin} presents the F1-Score, Average Precision, Area Under the Receiver Operating Characteristic and Sensitivity (detailed later in Section~\ref{metrics}) for the data mentioned above through application of three very popular anomaly detection techniques (Isolation Forest, Local Outlier Factor, One-Class Support Vector Machine) after a projection on a finite-dimensional subspace defined by means of the Haar basis. Although the results are barely satisfactory at first glance, in view of the high complexity of the data, the greater performance of methods targeting directly the functional nature of the data will be demonstrated in the analysis carried out in Section~\ref{bench}. While a fine tuning of the employed dimension reduction technique may possibly improve the result, this is far from simple in practice, insofar as it could require to implement an extremely large number of methods with many parameter configurations.

{\renewcommand{\arraystretch}{1.5}
{\setlength{\tabcolsep}{0.25cm}
\begin{table}[!h]
\begin{center}
{\scriptsize
\begin{tabular}{ccccc}
\hline
 Method & F1-Score &  AP & AUC& $p_c$ \\ \hline
 \multicolumn{5}{|c|}{ Aeronautics dataset } \\ \hline
IF &0.71&0.69& 0.49&0.27\\
LOF &0.68&0.73&0.5&0.21\\
OCSVM &0.78&0.77&0.63&0.45\\ \hline
\multicolumn{5}{|c|}{ Rocks dataset } \\ \hline
IF &0.971&0.978& 0.528&0\\
LOF &0.973&0.965&0.495&0.05\\
OCSVM &0.971&0.965&0.686&0\\ \hline
\end{tabular}}
\end{center}
\vspace*{0.2cm}
\caption{Finite-dimensional algorithms on Haar-basis projection coefficients  considered in performance comparison with F1-Score, Average Precision (AP), AUC and Sensitivity $(p_c)$ for both datasets.}
\label{tab:benchfin}
\end{table}}}

\subsection{Coping with the Functional Nature of the Data}
\label{func}

We now review recent functional depth based techniques, which may offer attractive practical alternatives to the traditional techniques previously recalled.
Suppose that the random variable of interest $\bmX$ modelling the normal behavior of the system monitored takes its values in a Hilbert space, say the space $L^2([0,1])$ of square integrable functions on $[0,1]$ for simplicity. Direct extension of multivariate data depth methods to this functional setting turns to be impractical because the resulting depth functions then vanish everywhere during the optimization, as pointed out in \textit{e.g.} \cite{KuelbsZ15}, due to the richness of the feature space. Further, a set of desirable properties is imposed on the defined functional depth statistic, which guarantees its usefulness in applications, see  \cite{nieto,gijbels2017general} for their overview. To adjust for these requirements, various strategies have been proposed. In particular, the search of the depth function is restricted to features from a dictionary of interest only in \cite{MoslerP18}, while integrating univariate depth functions over time is suggested in \textit{e.g.} \cite{claeskens2014multivariate} or \cite{hubert2015multivariate}. This second approach is often preferred in practice, due to its simplicity and (often substantially) lower computational burden.

\noindent  {\bf  Integrated functional depths.}  Let $\text{D}^{(1)} (\cdot \mid \cdot)$ be a univariate depth measure, refer to \textit{e.g.} \cite{ZuoS00} and~\cite{Mosler13} for an account of depth statistics. A typical example is the Tukey depth, see \cite{tukey1975mathematics} and also \cite{fraiman2001trimmed} or \cite{claeskens2014multivariate}. For an i.i.d. sample $\mathcal{D}_n=\{X_1,\; \ldots,\; X_n\}$ of real valued observations, it is defined as:
\begin{equation}\label{equ:univTukey}
	\text{D}^{(1)}_T(x\, \mid \, \mathcal{D}_n) = \min\{\hat{F}_{n}(x),1 - \hat{F}_{n}(x^-)\}\,,
\end{equation}
where the c\`ad-l\`ag function $\hat F_{n}(x)=(1/n)\sum_{i=1}^n\mathbb{I}\{X_i\leq x \}$ is the empirical cdf based on the dataset $\mathcal{D}_n$, denoting by $\mathbb{I}\{\mathcal{E}\}$ the indicator function of any event $\mathcal{E}$. Now, given a sample of curves $\mathcal{C}_n=\{\bmX_1,...,\bmX_n\}$ composed of independent copies of the functional observation $\bmX(t)$ with $t\in[0,1]$ valued in $L_2([0,1])$, the integrated functional data depth function defined below provides a centrality measure on the infinite dimensional space $L^2([0,1])$:

\begin{equation}\label{equ:depthInt}
	\text{D}(\bmX \mid \mathcal{C}_n) = \int_{0}^{1}\text{D}^{(1)}\bigl(\bmX(t)\, \mid \,\mathcal{C}_n(t)\bigr)dt,
\end{equation}
where $\mathcal{C}_n(t)=\{\bmX_1(t),...,\bmX_n(t)\}$ is the stamp of the functional sample at time point $t$.  While being purely data-driven and robust by construction, the univariate data depth \eqref{equ:univTukey} is non-continuous, and equals $0$ outside the range of the $X_i$'s. Another option is thus to consider the integrated functional depth based on the projection depth introduced in \cite{ZuoS00}, which is given by:
\begin{equation*}
	\text{D}^{(1)}_P(x\,\mid \, \mathcal{D}_n) = \left(1 + \frac{\lvert x-\text{med}(X) \rvert}{\text{MAD}(X)}\right)^{-1}\,,
\end{equation*}
where $\text{med}(X)$ and $\text{MAD}(X)$ are the median and the median absolute deviation based on the univariate sample $\mathcal{D}_n$. While being positive everywhere, projection depth is symmetric around the median. More generally, the asymmetric version proposed by \cite{hubert2015multivariate} is given by
\begin{equation*}
	\text{D}^{(1)}_{AP}(x\, \mid \,\mathcal{D}_n) = \frac{1}{1 + \text{AO}(x\, \mid \,\mathcal{D}_n)}\,,
\end{equation*}
where the asymmetric outlyingness $\text{AO}$ is defined as:
\begin{equation*}
	\text{AO}(x\,\mid \,\mathcal{D}_n) = \begin{cases}
		\frac{x - \text{med}(X)}{w_2(X) - \text{med}(X)} & \text{ if } x > \text{med}(X), \\
		\frac{\text{med}(X) - x}{\text{med}(X) - w_1(X)} & \text{ if } x < \text{med}(X)
	\end{cases}
\end{equation*}
with
\vspace*{-0.3cm}
\begin{align*}
	w_1(X) & = \hat q_X(0.25) - IQR(X) 1.5e^{-4\text{MC}(X)} \\
	w_2(X) & = \hat q_X(0.75) + IQR(X) 1.5e^{3\text{MC}(X)}\,,
\end{align*}
where $\text{IQR}(X)=\hat q_X(0.75) - \hat q_X(0.25)$ is the interquantile range with $\hat q_X$ being the empirical quantile based on $\mathcal{D}_n$ and $\text{MC}(X)$ is the robust measure of skewness proposed by \cite{BrysHS04} defined as follows:
\begin{equation*}
 \text{med}\left(\frac{\bigl(X_j + X_i - 2~\text{med}(X)\bigr) }{x_j - x_i}:  (i,j)\in \mathcal{I}\right),
\end{equation*}
where $\mathcal{I}=\{(i,j):\, i\neq j,\, X_i\le\text{med}(X)\le X_j \}$. 
In the subsequent empirical analysis, three integrated functional data depth functions are considered, calculated as the average Tukey depth (fT), projection depth (fSDO) and asymmetric projection depth (fAO) over the definition domain, respectively.

\begin{figure}[!h]
\begin{multicols}{2}
 \includegraphics[scale=0.3,trim=1.9cm 1cm 0 0cm,clip=true]{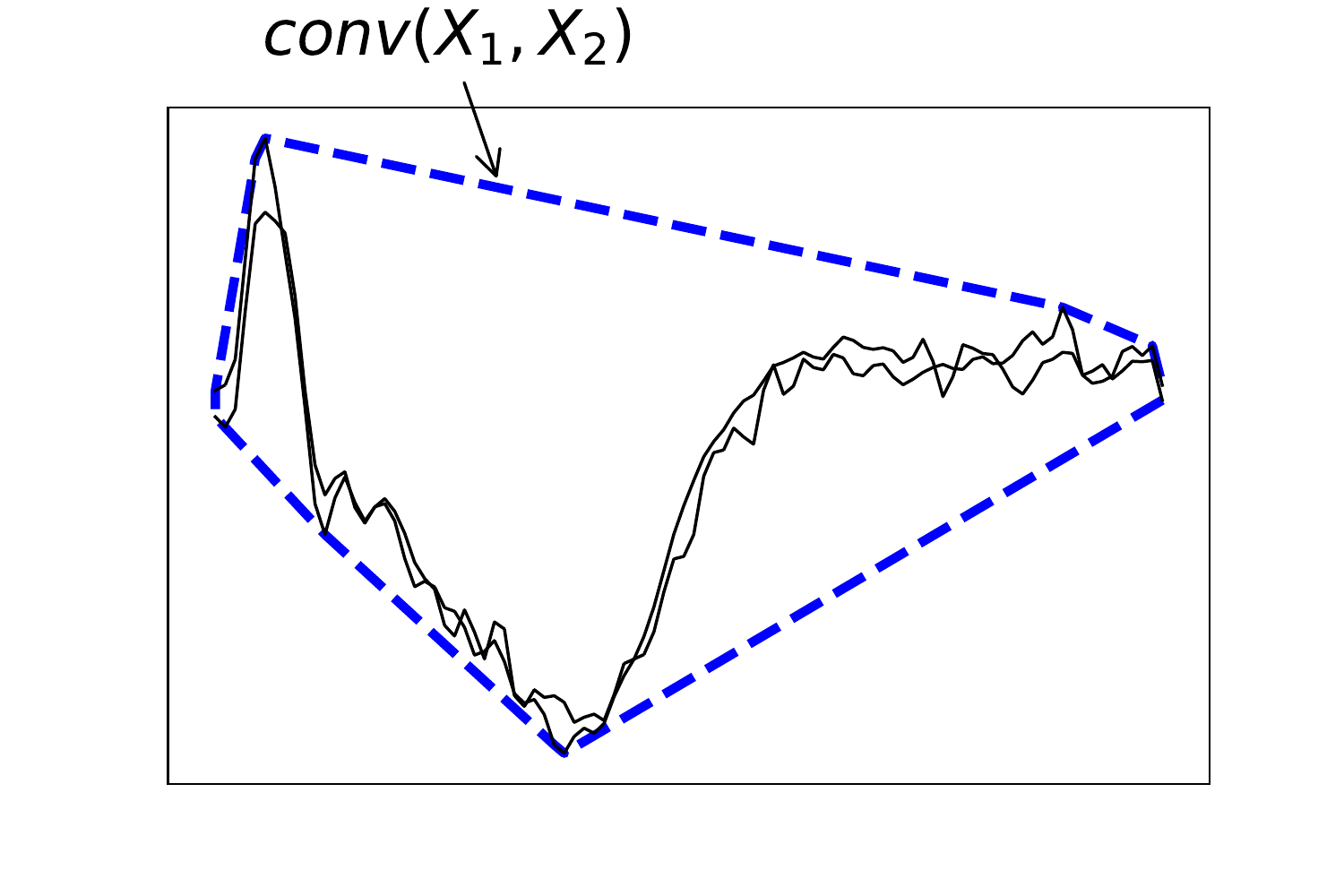} \\ \includegraphics[scale=0.3,trim=1.9cm 1cm 0 0cm,clip=true]{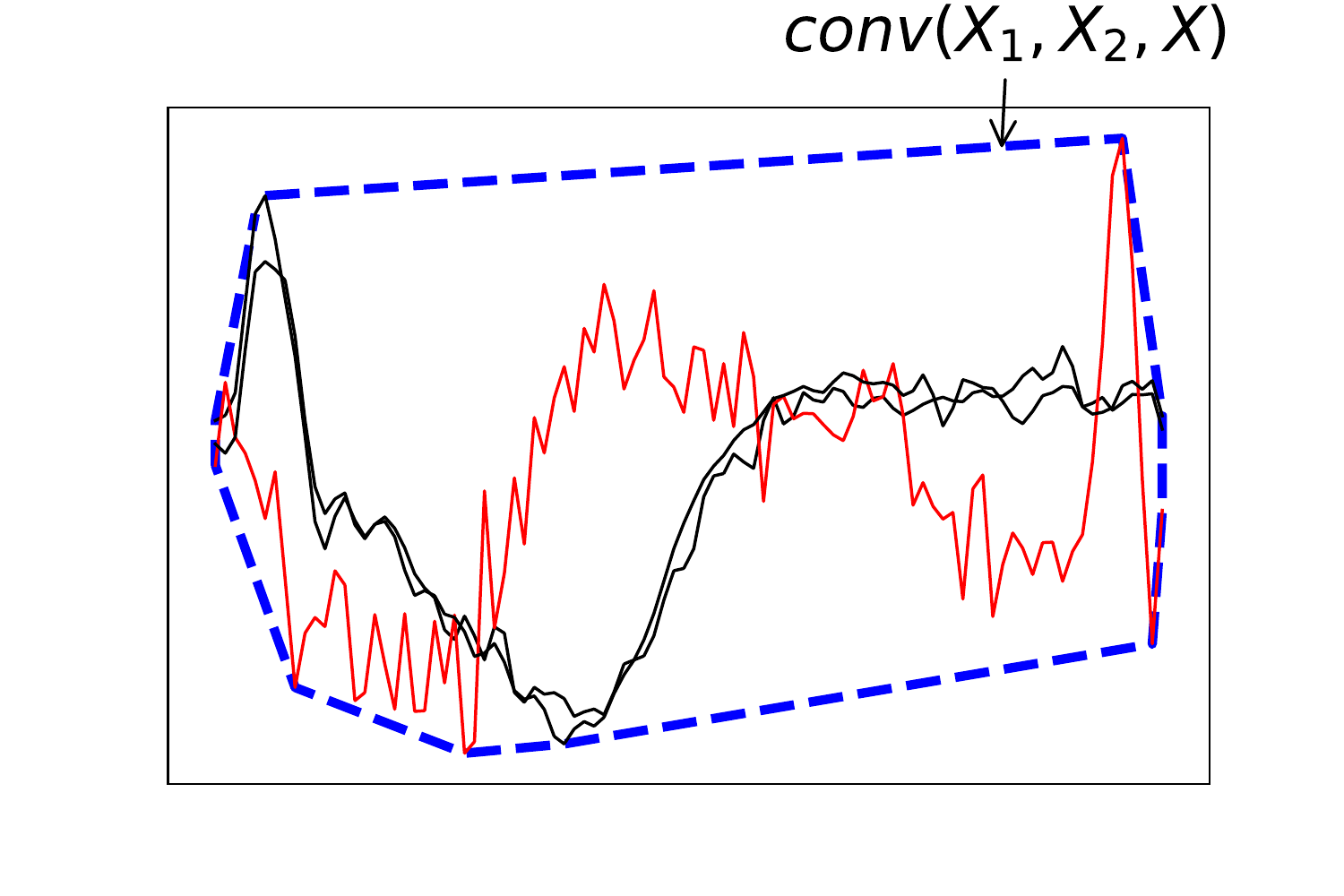} \\
\indent\\ \includegraphics[scale=0.3,trim=1.9cm -1cm 0cm 1.2cm,clip=true]{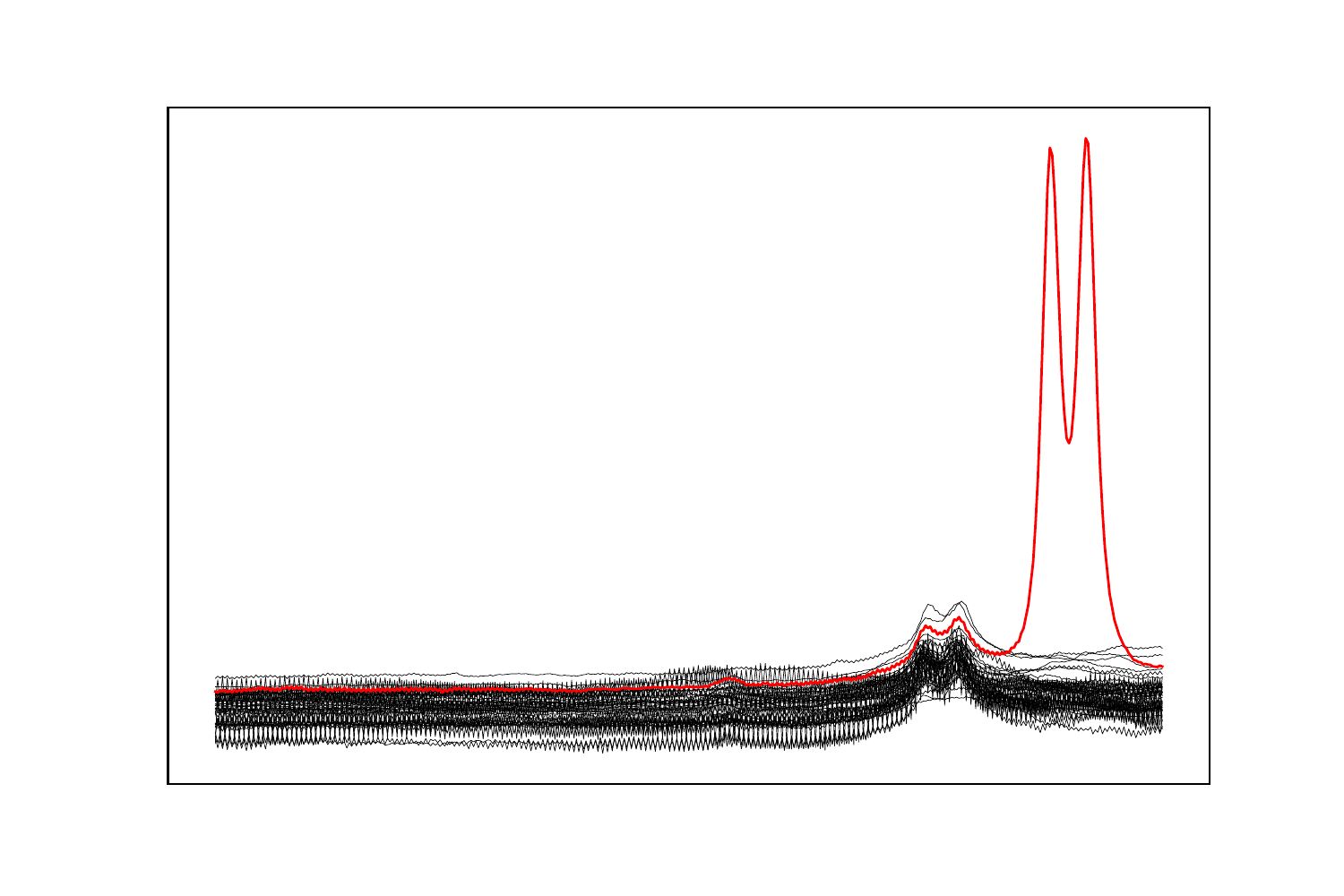} \\ \includegraphics[scale=0.3,trim=3cm 0cm 0 2cm,clip=true]{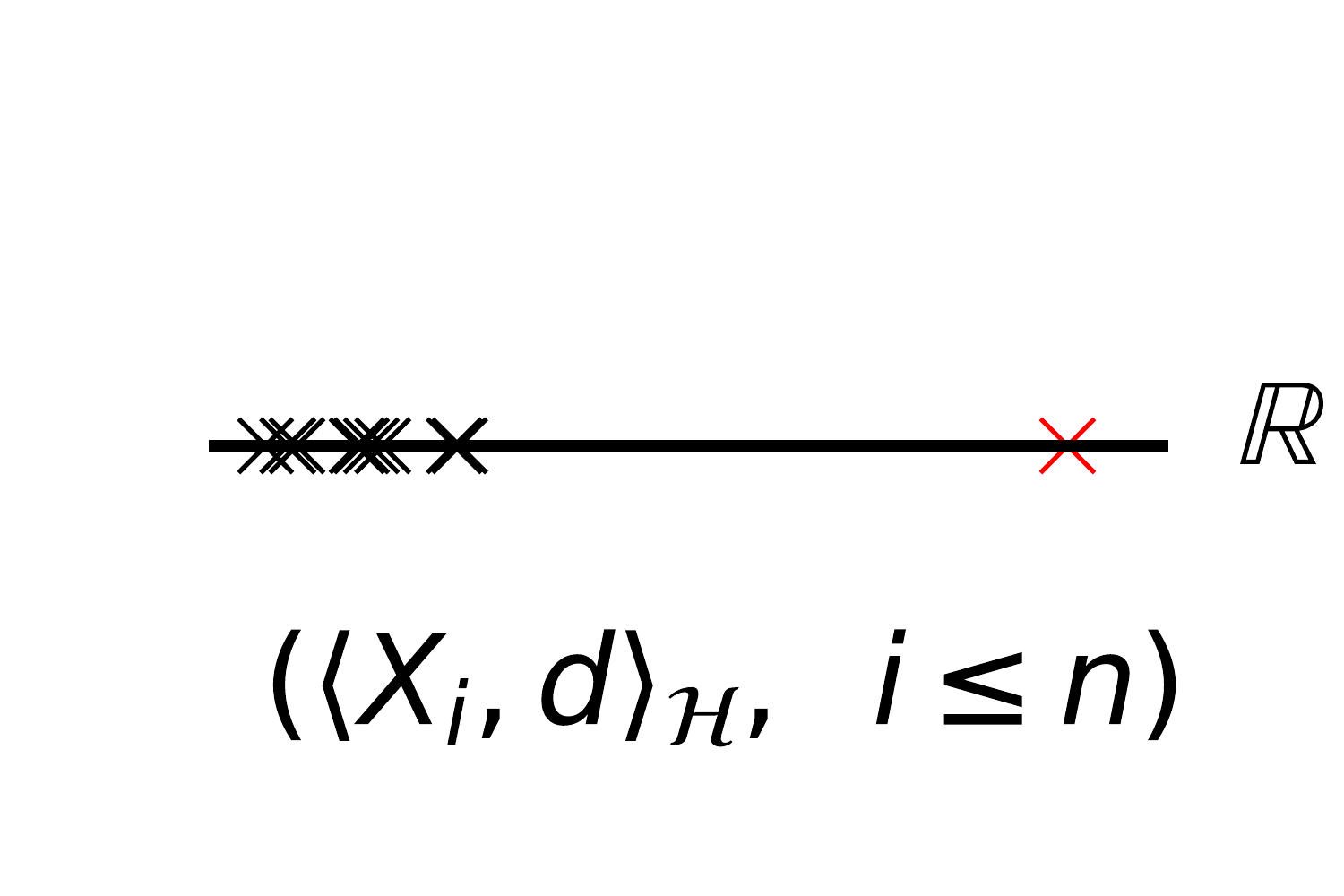}
\vspace*{-6cm}
\end{multicols}
\caption{Illustration of the influence of the convex hull of a single function in the ACH depth (left, top and bottom, abnormal (red) observation increases the volume) and separation in the scalar product in FIF (right, top and bottom, abnormal (red) observation lies far from the normal data).}\label{achfif}
\end{figure}

\noindent {\bf The Area of the Convex Hull (ACH) depth.} The area of the convex hull (ACH) depth, recently introduced by~\cite{staerman2019area}, is also considered in the present study. We point out that this functional depth function does not belong to the family of integrated depths \eqref{equ:depthInt} and exhibits high sensitivity beyond the convex envelope of the data. In short, the ACH depth quantifies the contribution of a given curve to the spread of the convex hull of a random (sub)sample of curves, see Figure~\ref{achfif} left. For a chosen degree $1 \le J\le n$, the ACH depth, denoted by $\text{D}_{ACH}(\bmX \, \mid \,\mathcal{C}_n)$ here, is defined as follows:

\begin{equation*}
	{\small \frac{1}{\displaystyle\binom{n}{J}}\sum_{1\le i_1\le\dots\le i_j\le n}\frac{\lambda(\text{conv}(\text{graph}(\{\bmX_{i_1},\dots,\bmX_{i_j}\})))}{\lambda(\text{conv}(\text{graph}(\{\bmX_{i_1},\dots,\bmX_{i_j}\}\cup\{\bmX\})))}}\,,
\end{equation*}

\noindent where $\lambda$ means the Lebesgue measure on $\mathbb{R}^2$, $\text{conv}(A)$ is the convex hull of any subset $A$ of the plane and $\text{graph}(\{\bmX_1,\dots,\bmX_n\})=\bigcup_{i=1}^n \text{graph}(\bmX_i)$ with the notation $\text{graph}(\bmX_i)=\{(t,y): y=\bmX_i(t), t\in [0,1]\}$. In \cite{staerman2019area} empirical evidence that even small values of $J$ ($J=2$ for instance) are sufficient to define a robust and sensitive functional depth that is relevant for anomaly detection, the parameter $J$, when picked in this range, having little impact on the preorder induced by the resulting depth function. One may refer to \cite{staerman2019area} for a detailed description of this approach.
\smallskip

\noindent {\bf Functional Isolation Forest.} Recently, \cite{staerman2019functional} extended the popular Isolation Forest approach, see \cite{LiuTZ08}, to the functional framework. A Functional Isolation Forest (FIF) is formed by a collection of functional isolation trees (F-$i$trees), constructed each (by a sequence of random splits) from a subsample (of size $\psi$) of $\mathcal{C}_n$. The abnormality score of the observation $\bmX$ is then computed as a monotone decreasing transformation of the average depth of $\bmX$ over the trees. The main idea here is that, since the splits are purely random, a very differing observation will be cut-off (isolated) from $\mathcal{C}_n$ with higher probability (and thus on less deep levels of the F-$i$trees) than those similar to the majority of the observed curves. The construction of the F-$i$trees is based on a pre-determined dictionary $\mathbf{D}$ that can contain both deterministic and/or stochastic functions that capture relevant properties of the data, which can also be a subset of $\mathcal{C}_n$. Before each random univariate split, all observations of the F-$i$tree are projected on a line spanned by a random element of $\mathcal{D}$, see Figure~\ref{achfif} right. The choice of a suited dictionary thus plays a crucial role in the construction of the FIF score. This projection is calculated by means of the scalar product designed to account for both location and shape anomalies, \textit{i.e.} for any $\alpha\in[0,1]$: 
\begin{equation*}\label{eq:scalarprodFif}
\langle \bmX, \bmd\rangle :=\alpha\times\frac{\langle \bmX, \bmd\rangle_{L_2}}{\|\bmX\|\|\bmd\|}+(1-\alpha)\times\frac{\langle \bmX', \bmd'\rangle_{L_2}}{\|\bmX'\|\|\bmd'\|}, 
\end{equation*}
where $\bmd\in\mathbf{D}$, $\bmX'$ and $\|\bmX\|$ are respectively the first derivative and the $L_2$-norm of  $\bmX$ in $L_2([0,1])$. 

\section{A Preparatory Simulation Study}\label{sec:simulation}

This section is devoted to empirical analysis of the performance of the functional anomaly detection techniques whose mechanics have been briefly described in Section~\ref{sec:methods}. Their accuracy is investigated from simulated data inspired from a real dataset collected by Airbus, composed of one-minute sequences of accelerometer data measured on helicopters. As a first go, we recall the standard performance metrics commonly used.



\subsection{Performance Metrics in Anomaly Detection}\label{metrics}

Although the learning procedure does not rely on data of the same nature, the anomaly detection problem can be formulated in the same probabilistic framework as binary classification, the flagship problem in statistical learning. In the standard setup, the binary random variable $Y$ indicates the occurrence of an anomaly: the label is positive, \textit{i.e.} $Y=+1$, when an anomaly occurs, and negative, \textit{i.e.} $Y=-1$, otherwise. The random variable $\bmX$, taking its values in the feature space $\mathcal{X}$, models the measurement at disposal to predict $Y$. The goal pursued is to build an anomaly scoring function $s:\mathcal{X}\rightarrow \mathbb{R}\cup\{+\infty\}$ in an unsupervised manner (without observing $Y$), so that, ideally, the larger $s(\bmX)$, the likelier an anomaly occurs, \textit{i.e.} the more probable is $Y=+1$. A decision to raise an alarm can be then built by thresholding the scoring function $s(\bmx)$ at a critical level, ruling the trade-off between errors of type I and type II.  Equipped with this notation, decreasing transforms of a depth function w.r.t. $\bmX$'s marginal distribution provide anomaly scoring functions in a natural fashion. Precisely,  when using data depths in the following study,  the transformation $1-D(.)$ is performed to rescale them as an anomaly score.
\smallskip

\noindent {\bf ROC analysis.} The golden standard to quantify theoretically the accuracy of an anomaly scoring function $s$ is the PP-plot of the false positive rate \textit{vs} the true positive rate, namely $t\in\mathbb{R}\mapsto (\mathbb{P}\{s(\bmX)\geq t \mid Y=-1 \},\; \mathbb{P}\{s(\bmX)\geq t \mid Y=+1 \})$, referred to as the ROC curve (standing for Receiver Operator Characteristic curve), see \textit{e.g.} \cite{Fawcett06}. The higher the curve, the more accurate the anomaly scoring function. A simple Neyman-Pearson argument shows that optimal scoring functions are increasing transforms of the likelihood ratio $\Psi(\bmX):=(dF_+/dF_-)(\bmX)$, denoting by $F_{\sigma}$ the conditional distribution of $\bmX$ given $Y=\sigma1$, $\sigma\in\{-,\; +\}$: their ROC curve dominating everywhere the ROC curve of any other anomaly scoring function. For this reason, this functional performance measure is generally summarized by the Area Under the ROC curve (AUC in abbreviated form), a popular scalar criterion that can be classically interpreted as the rate of concordance of pairs: $\mathrm{AUC}(s)=\mathbb{P}\{ s(\bmX)>s(\bmX')\mid Y=+1,\; Y'=-1 \}+\mathbb{P}\{ s(\bmX)=s(\bmX')\mid Y=+1,\; Y'=-1 \}/2$, where $(\bmX',Y')$ denotes an independent copy of the pair $(\bmX,Y)$.
\smallskip

\noindent {\bf PR analysis.} Alternatively, one may evalute the accuracy of any score by plotting the precision-recall (shortly PR) curve, namely $t\mapsto (\mathbb{P}\{s(\bmX)\geq t \mid Y=+1 \},\; \mathbb{P}\{Y=+1 \mid s(\bmX)\geq t \}) $. The higher its PR curve, the more accurate an anomaly scoring function. Of course, as may be immediately shown by means of the Bayes formula, PR and ROC curves are in one to one correspondence, see \textit{e.g.} \cite{CV11}. Like the ROC curve, the PR curve may be summarized by the area under it, referred to as the Average Precision (AP).

When labeled data are available, it is possible to compute the performance measures recalled above, replacing the probabilities involved by their statistical counterparts, in order to assess the accuracy of any anomaly scoring function candidate $s(\bmx)$. However, in unsupervised anomaly detection, the scoring function $s(\bmx)$ cannot be learned using labeled training data, in contrast to classification or bipartite ranking: only 'negative' observations, \textit{i.e.} an i.i.d. sample drawn from (a possibly noisy version of) distribution $F_-$, are available in the training stage. Hence, the learning task cannot be achieved by optimizing empirical versions of the aforementioned criteria, which makes it extremely challenging.

\subsection{Simulating Anomalies of Specific Types}\label{sim}

\begin{figure}[!h]
\begin{center}
\begin{tabular}{cc}
\hspace*{0.3cm}  Isolated & \hspace*{0.3cm}  Magnitude I \\
\includegraphics[scale=0.4,trim=0 0 1cm 0.75cm,clip=true]{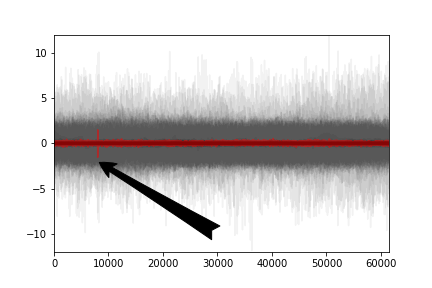} & \includegraphics[scale=0.4,trim=0 0 1cm 0.75cm,clip=true]{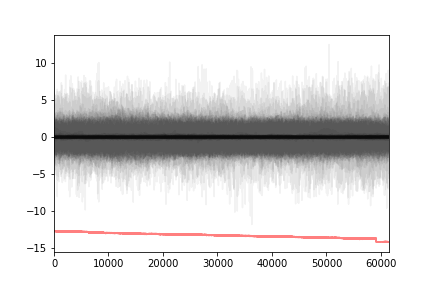} \\
\hspace*{0.3cm}  Magnitude II & \hspace*{0.3cm}  Shape \\
\includegraphics[scale=0.4,trim=0 0 1cm 0.75cm,clip=true]{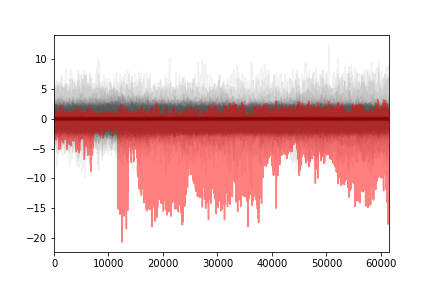} & \includegraphics[scale=0.4,trim=0 0 1cm 0.75cm,clip=true]{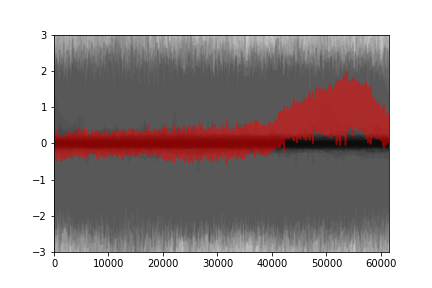} \\
\end{tabular}
\end{center}
\caption{Examples for each of the identified types of anomalies in the aeronautics dataset. In order from top to bottom and from left to right: each time one isolated, magnitude, magnitude/shape and shape anomaly, with a subsample of normal data. Grey curves are normal data while red curves are anomalies.}
\label{anom_data}
\end{figure}

Datasets containing various types of anomalies are usually more challenging to analyze. As a first go, we start by investigating to which extent the techniques recalled above may permit to detect simulated anomalies of well-identified types according to the usual taxonomy, see, \textit{e.g.}, \cite{hubert2015multivariate}. Figure~\ref{anom_data} illustrates the four types of anomalies addressed in detail in these experiments: isolated, magnitude (of two different kinds) and shape anomalies.

To reproduce a controlled version of each type of anomalies, four datasets have been built from a collection of $1794$ 'normal' functional observations from the validation dataset collected by Airbus. Each functional observation corresponds to accelerometer data measured on helicopters at a $1024$ Hz frequency over time windows of $1$ minute: the curves $\bmX=(\bmX(t))_{t\in[0,1]}$ are built by means of an affine interpolation of the $61 440$ sampled points. One per anomaly type, four datasets have been constructed by adding a specific contamination to $5\%$ of these 'normal' observations, drawn uniformly at random. Cases when $1\%$, $2\%$ $3\%$ and $4\%$ are added in the Appendix \ref{additional} for completeness and show similar behavior of methods than for $5\%$. The four contamination models defined below, are used to generate independent curves $\bmY$ (independently from the original dataset) that are next added to the selected above $1794$ 'normal' observations $\bmX$. By $\mathcal{U}([a,b])$ is meant the uniform distribution on the interval $[a,b]$, while $\delta_u$ denotes the Dirac mass at point $u$.
\smallskip

\noindent \textbf{Model 1 (Isolated Anomalies)}\\
\noindent $\bmY(t)=\varepsilon u_1\mathbb{I}\{t=\tau\} $, where  $u_1 \sim \mathcal{U}([3,4])$ and $\varepsilon \sim (1/2)(\delta_{-1}+\delta_1)$ are independent random variables, with $\tau$ being the time at which the isolated anomaly occurs that is chosen randomly in a uniform manner among the set of sampling points, independently of $u_1$ and $\varepsilon$.
\smallskip

\noindent \textbf{Model 2 (Magnitude Anomalies I)}

\noindent $\bmY(t)\equiv u_2$, with $u_2\sim \mathcal{U}([-12,-15])$.
\smallskip

\noindent \textbf{Model 3 (Magnitude Anomalies II)}

\noindent  $\bmY(t)= u_3 \mathbb{I}(t\in I)$, where $u_3 \sim \mathcal{U}([0,15])$ and $I$ is a subinterval of $[0,1]$ of length $1/10$ whose location is chosen uniformly at random, independently of $u_3$.
\smallskip

\noindent \textbf{Model 4 (Shape Anomalies)}

\noindent $\bmY(t)= \sin(2\pi u_4 t)$, where $u_4 \sim \mathcal{U}([0.2,2])$.

\begin{figure}[!h]
\begin{center}
\begin{tabular}{cc}
\hspace*{0.3cm} Isolated &  \hspace*{0.3cm}  Magnitude I \\
\includegraphics[scale=0.4,trim=0 0 1cm 0.75cm,clip=true]{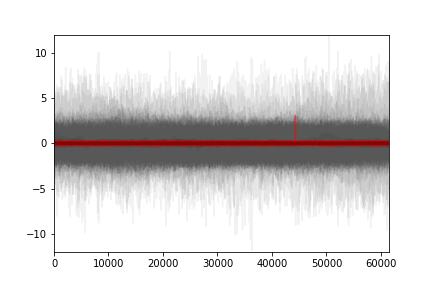} & \includegraphics[scale=0.4,trim=0 0 1cm 0.75cm,clip=true]{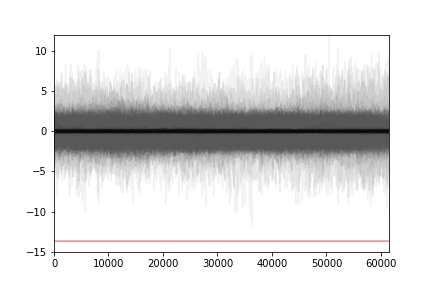} \\
\hspace*{0.3cm}  Magnitude II & \hspace*{0.3cm}  Shape \\
\includegraphics[scale=0.4,trim=0 0 1cm 0.75cm,clip=true]{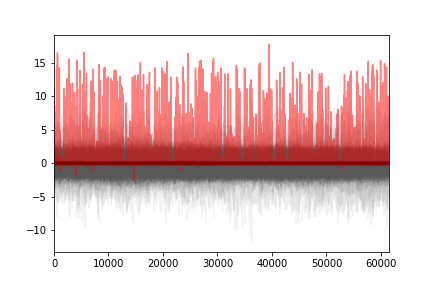} & \includegraphics[scale=0.4,trim=0 0 1cm 0.75cm,clip=true]{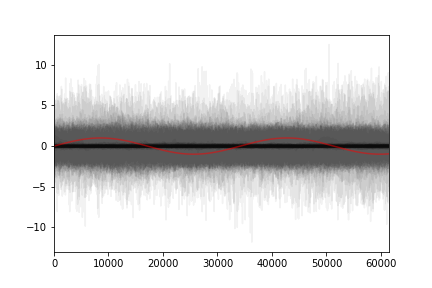} \\
\end{tabular}
\end{center}
\caption{Examples for each of four types of simulated anomalies. In order from top to bottom and from left to right: each time one isolated, magnitude, magnitude/shape and shape anomaly, with a subsample of normal curves (out of 1794). Grey curves are normal data while red curves are simulated anomalies.}\label{simanom_data}
\end{figure}

Simulated anomalies, together with a small subset of 'normal' data, are illustrated in Figure~\ref{simanom_data}.

{\renewcommand{\arraystretch}{1.1} 
{\setlength{\tabcolsep}{0.2cm}
\begin{table}[!h]
\begin{center}
{\scriptsize
\begin{tabular}{p{2cm}|cccc}
\hline 
Anomaly type & & IF& LOF& OCSVM\\ \hline
\multirow{2}{*}{\textbf{Isolated} }& $p_c$ &0 &0&0\\
& AUC&0.41&0.25&0.44\\ \hline
\multirow{2}{*}{\textbf{Magnitude I}} &$p_c$  &1 &0.48&1  \\
&AUROC&1& 0.97& 1 \\\hline
\multirow{2}{*}{\textbf{Magnitude II} }&$p_c$ &0&0&0\\
&AUC& 0.54& 0.03& 0.7\\\hline
\multirow{2}{*}{\textbf{Shape} } &$p_c$ &0 &0.48& 0 \\
& AUC&0.67 & 0.97& 0.67 \\\hline
\end{tabular}}
\end{center}
\vspace*{0.2cm}
\caption{Methods considered in performance comparison with the sensitivity $(p_c)$ and the Area Under the Receiver Operating Characteristic (AUROC) for the four simulated models.}
\label{tab:simu}
\end{table}}}

\onecolumn

{\renewcommand{\arraystretch}{1.5} 
{\setlength{\tabcolsep}{0.4cm}
\begin{table}[!h]
\begin{center}
{\scriptsize
\begin{tabular}{p{2.5cm}cccccccc}
\hline
& \multicolumn{2}{c}{\textbf{Isolated} }&\multicolumn{2}{c}{\textbf{Magnitude I}}&\multicolumn{2}{c}{\textbf{Magnitude II} }&\multicolumn{2}{c}{\textbf{Shape}} \\
Methods & $p_c$ &AUC&$p_c$ &AUC&$p_c$ &AUC&$p_c$ &AUC\\ \hline
FIF &0 &0.21&0.93&0.99&0&0.30&\textbf{0.63}&\textbf{0.98}\\
fAO &0 &0.44&\textbf{1}&\textbf{1}&0&0.54&0&0.66\\
fbd &0 &0.44&\textbf{1}&\textbf{1}&0&0.54&0&0.68\\
fSDO & 0&0.42&\textbf{1}&\textbf{1}&0&0.43&0&0.77\\
fT& 0&0.43 &\textbf{1}&\textbf{1}&0&0.44&0& 0.71 \\
ACH&0 & \textbf{0.63}&0.48&0.97&\textbf{0.80}&\textbf{0.99}&0& 0.56 \\
Outliergram&0 &0.55&\textbf{1}&\textbf{1}&0&0.54&0& 0.47 \\
MS + IF &0& 0.05&0.66 &0.98&0&0.70&0.33& 0.74 \\
FOM (fSDO) + IF &0 &0&0.85&0.99&0.64&0.99&0.06& 0.80 \\
FOM (fAO) + IF &0 &0.14&0.81 &0.99&0.55&0.98 &0.02&0.87 \\
FPCA + IF &0 &0.11&0&0.91&0& 0.71&0.45&0.97\\
FPCA + LOF &0&0.5& 0&0.16&0&0.38&0&0.79\\
FPCA + OC&0 &0.04&0&0.93&0&0.77&0.3&0.96\\ \hline
\end{tabular}}
\end{center}
\vspace*{0.2cm}
\caption{Methods considered in performance comparison with the sensitivity $(p_c)$ and the Area Under the Receiver Operating Characteristic (AUROC) for the four simulated models  with 5\% of added anomalies. }
\label{tab:simu2}
\end{table}}}
\begin{figure}[!h]\label{roc1}
\begin{center}
\begin{tabular}{cc}
Isolated & Magnitude I\\
\includegraphics[scale=0.5]{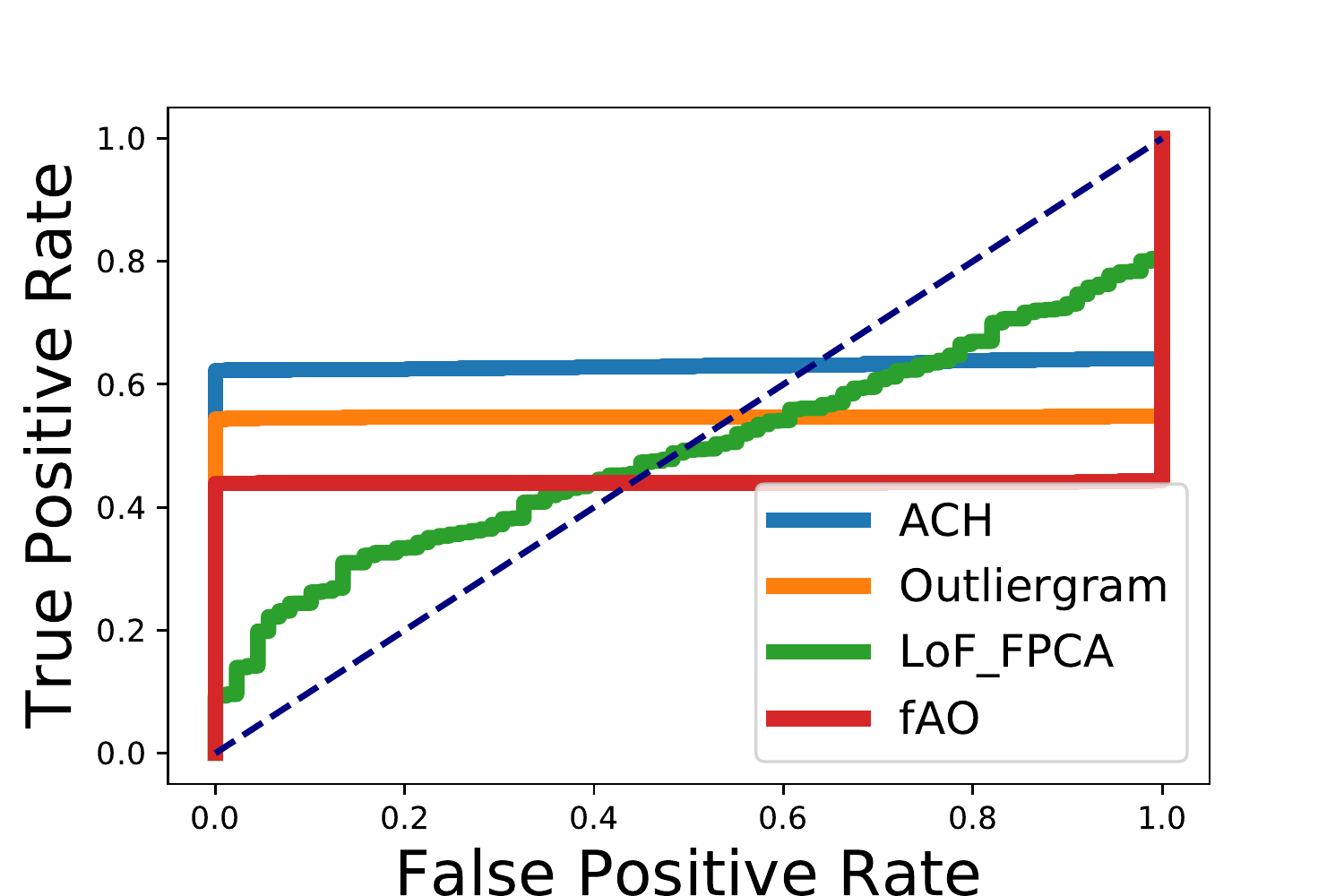}&\includegraphics[scale=0.5]{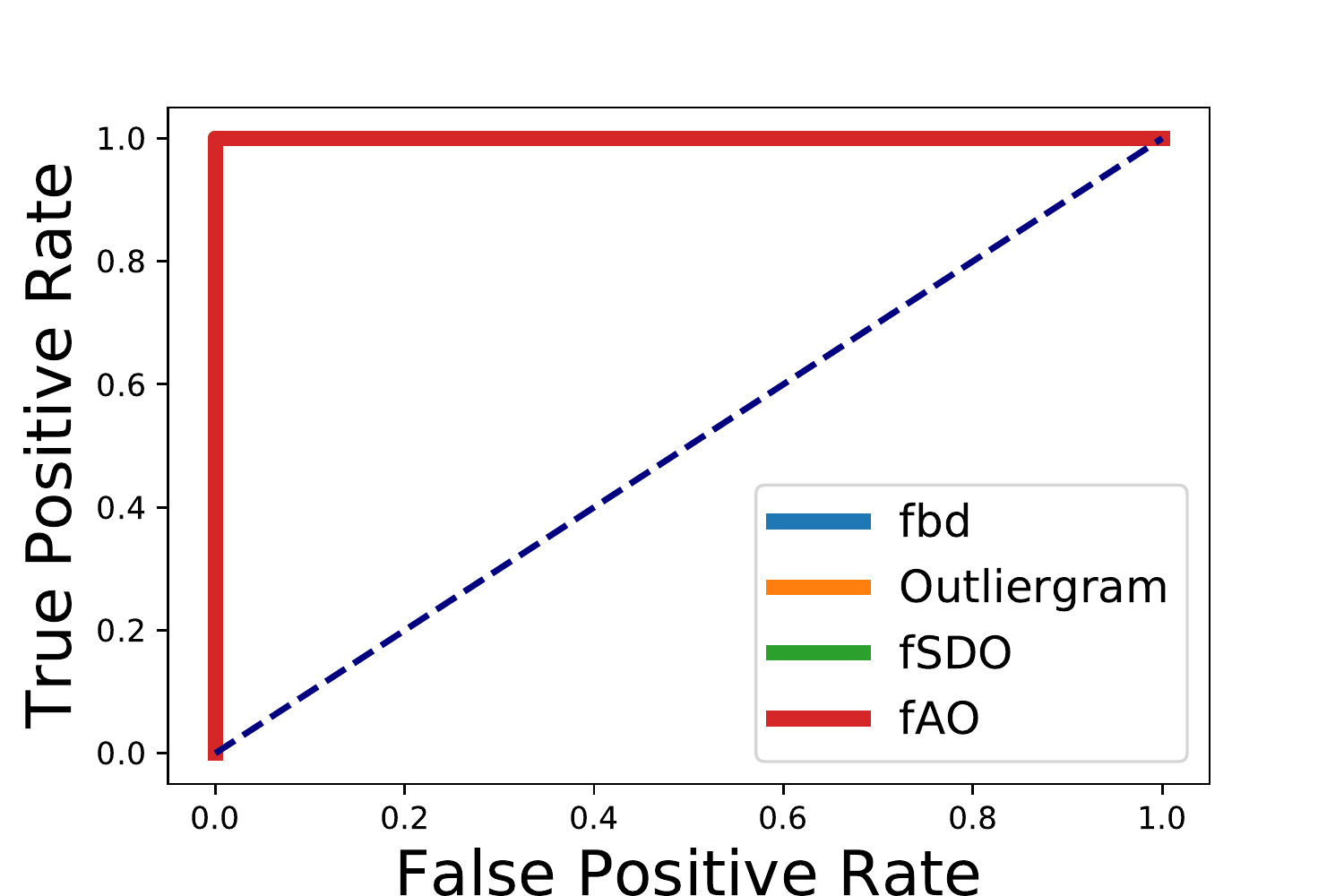}\vspace*{1cm}
\\

Magnitude II & Shape\\
\includegraphics[scale=0.5]{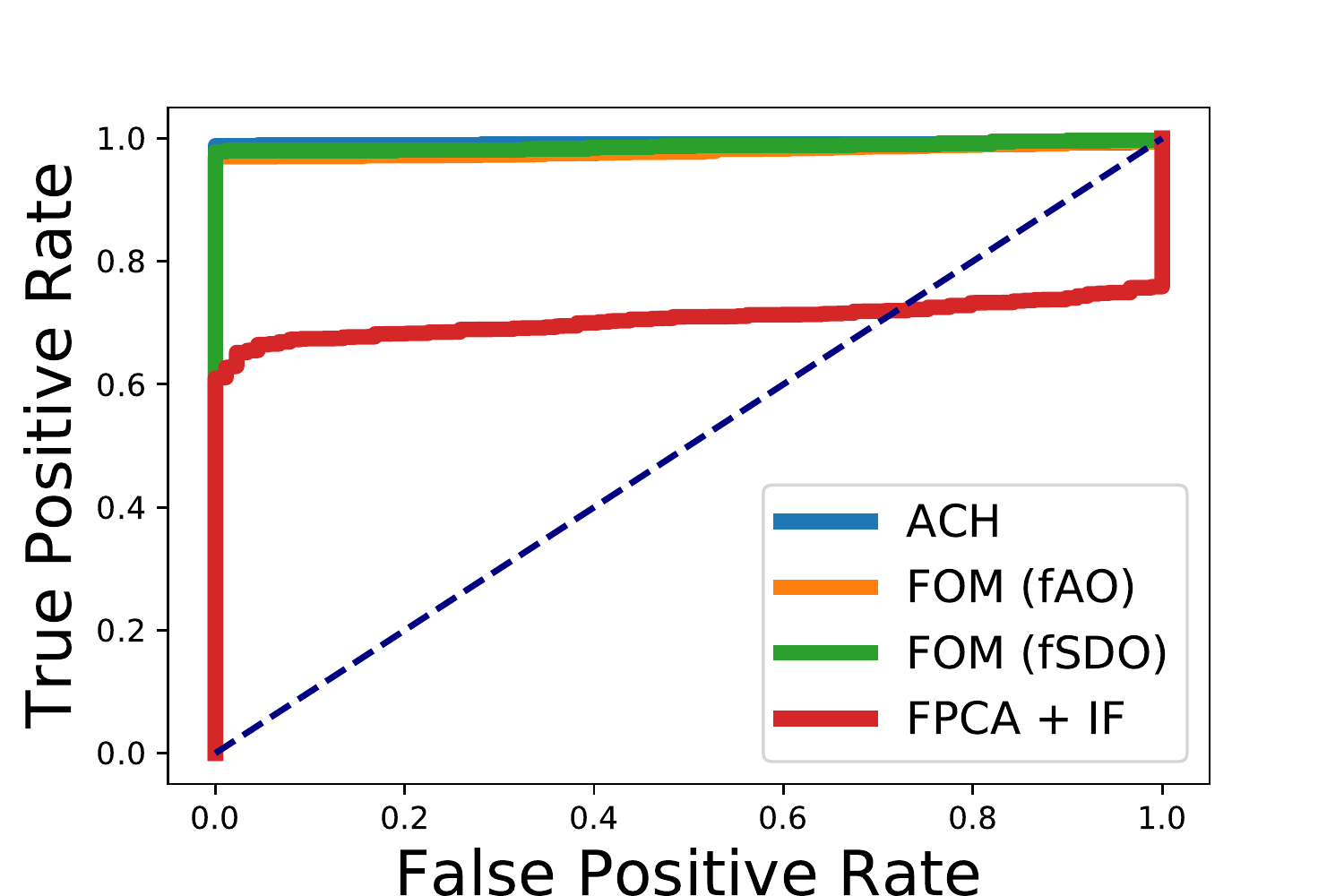}&\includegraphics[scale=0.5]{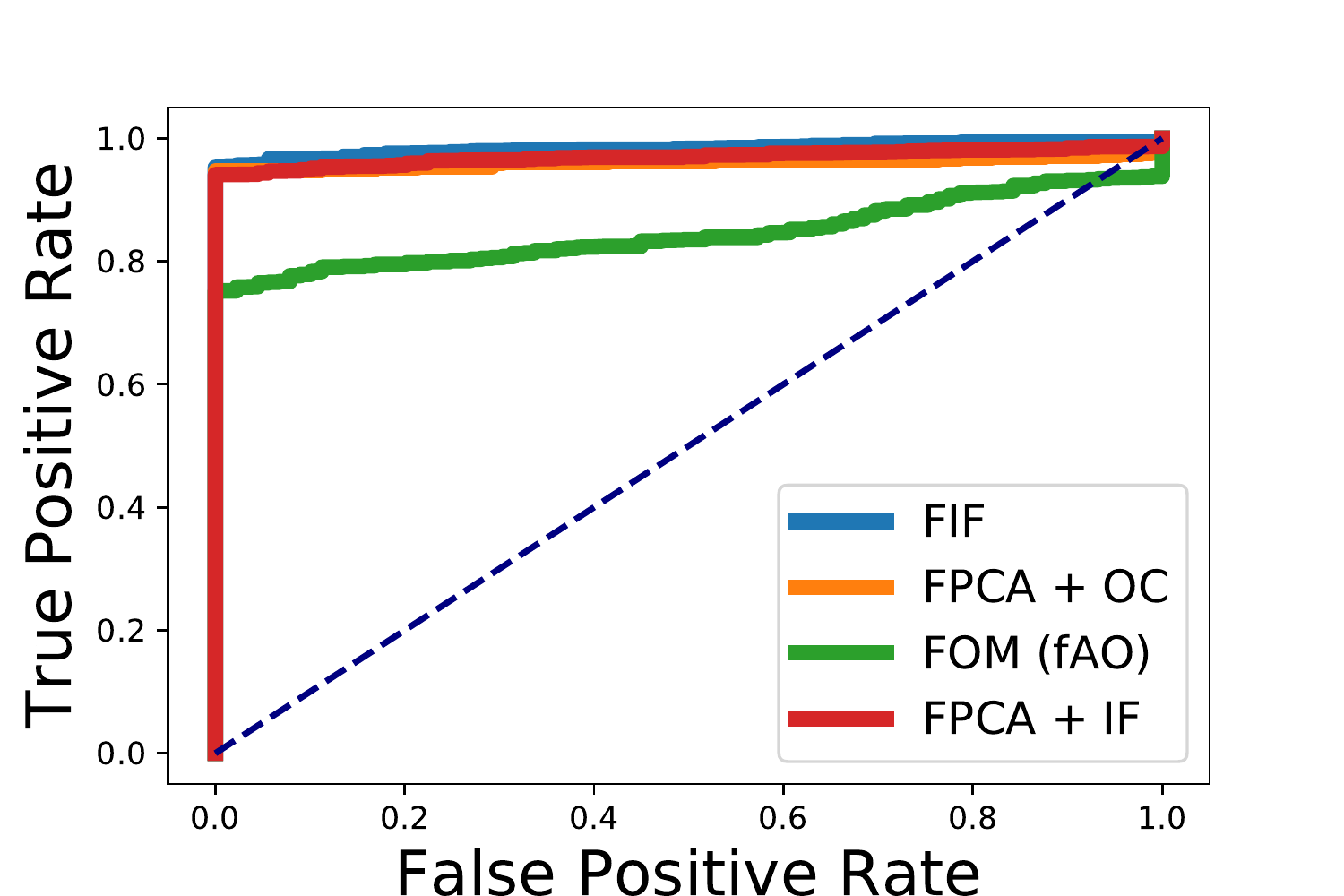}
\end{tabular}
\caption{ROC curves of the four best methods for the simulation study.}
\label{roc1}
\end{center}
\end{figure}
\twocolumn

\subsection{Naive Approaches - Sampled Curves Viewed as Multivariate Data}

Being the most straightforward idea, and still frequently employed in functional anomaly detection, direct consideration of (discretized) functional data in the multivariate space (\textit{i.e.}, $\mathbb{R}^p$) of their measurements can be seen as a first na\"{i}ve approach. It is important to notice that the sampling design should not vary with the curve/signal and preferably correspond to regularly spaced points in the observation domain, so that no dimensions are disadvantaged. If not, one can resort to importance-weighting techniques though, \textit{e.g.}, giving more weights to coordinates with higher marginal variance as it is suggested, \textit{e.g.}, in \cite{claeskens2014multivariate}. 
When implementing such a na\"{i}ve strategy to deal with functional observations, specific attention must be paid to the possibly very large dimension of the data (compared to the size of the population sample), due to the high frequency character of the measurements, as it is the case for the aeronautics dataset considered in this paper where $61440=p \gg n=1677$. We have applied three of the most widely used methods for multivariate anomaly detection, namely Isolation Forest (IF), Local Outlier Factor (LOF), and One-Class Support Vector Machine (OCSVM) to the four settings described in Section~\ref{sim}.

Their sensitivity and Area Under Receiver Operating Characteristic (AUC) are reported in Table~\ref{tab:simu}.

\subsection{Results and Discussion}\label{ssec:simurd}


We now consider the set of methods which are in the center of attention of the current work. These treat the data directly in their original functional space, and---as we shall see right below---prove beneficial for anomaly detection.

The performance of these unsupervised methods is evaluated on the four simulated datasets described in Section~\ref{sim} using the sensitivity (portion of correctly identified anomalies, $p_c$) and AUC as metrics. All parameters of the used algorithms are set to their default values (as it is pre-defined in the corresponding software packages). Thus, fSDO, fAO, fbd, FOM are available in the \texttt{mrfDepth} \texttt{R}-package~\cite{segaert}; the outliergram is available in the \texttt{roahd} \texttt{R}-package~\cite{tarabelloni}; IF, LOF and OC are available in \texttt{sklearn} \texttt{python} library~\cite{scikit-learn}; Magnitude-Shape plot (MS) is available in the \texttt{scikit-fda} \texttt{python} library~\footnote{\url{https://github.com/GAA-UAM/scikit-fda}}; Functional Isolation Forest (FIF) and ACH open-source python codes are available under the following link \footnote{\url{https://github.com/GuillaumeStaermanML}}; fT can be easily coded from scratch.\\

Results of the simulation study are displayed in Table~\ref{tab:simu2}. For completeness, the ROC curves of the four best methods for each contamination setting are displayed in Figure~\ref{roc1}.
As expected, the score drastically varies across contamination models and anomaly detection methods. Isolated anomalies (especially short ones) of the Contamination Model 1 are difficult to detect with projections on most bases as well as by integrating depths, whilst ACH is sensitive to this kind of anomalies. Magnitude (especially type I) anomalies are known to be easier to detect and a number of methods (fAO, fbd, fSDO, fT and outliergram) perform well by managing to detect all of them. The difficulty that differentiates Magnitude II anomalies (from those in Magnitude I) is that the anomalies are expressed only for a subset of time points. This impedes many methods from detecting this kind of anomalies, while ACH seems to perform best among differing results, most probably due to slight resemblance of Magnitude II anomalies with the isolated ones. Shape anomalies is the least identifiable type, and FIF delivers better performance than other methods while taking into account both location and slope of the functional curves (due to the employed Sobolev-type metric).


\section{Benchmarking Methods for Functional Anomaly Detection using Real Data}\label{sec:benchmark}

This section is devoted to the empirical analysis of the performance of the functional anomaly detection techniques whose mechanics have been briefly described in Section~\ref{sec:methods}. Their accuracy, previously investigated based on artificially contaminated data, shall be now benchmarked using real labeled datasets.

\onecolumn
\begin{figure}[!t]
\begin{center}
\begin{tabular}{cc}
\hspace*{0.3cm} FPCA &\hspace*{0.3cm}   MS-plot \\
\includegraphics[scale=0.4,trim=0 0 1cm 0.75cm,clip=true]{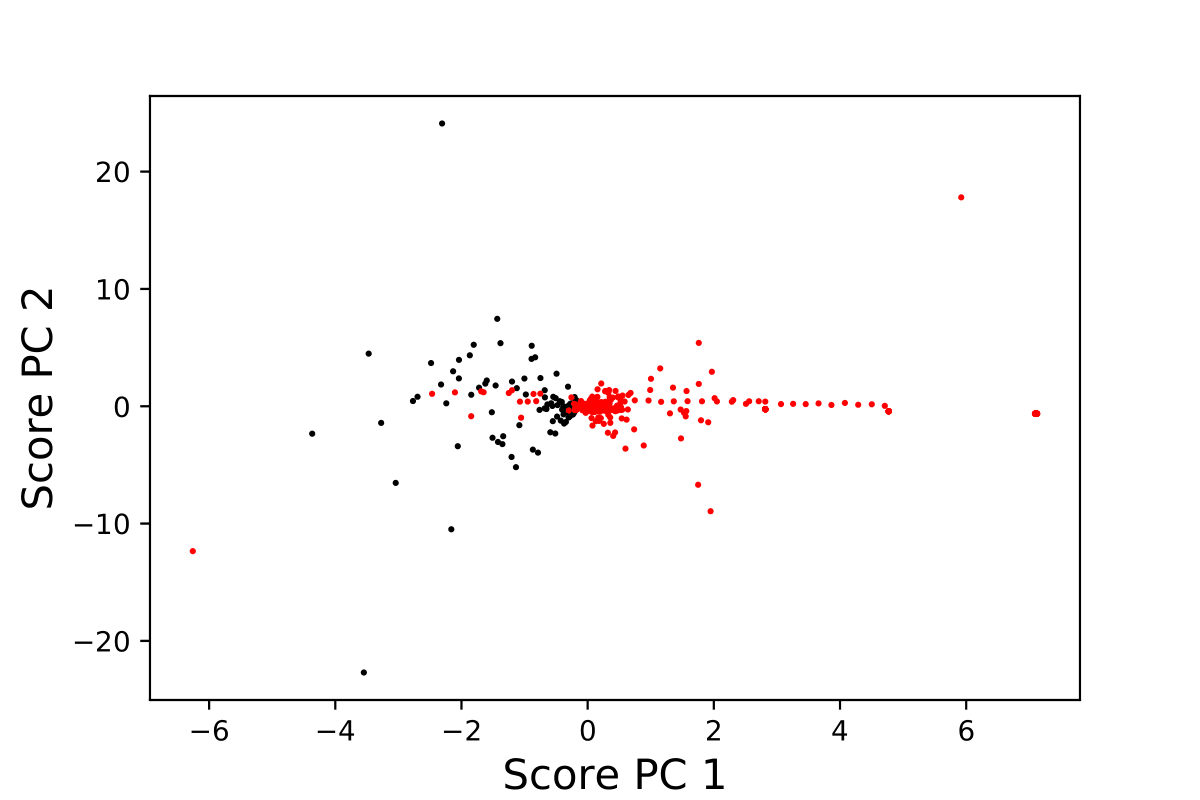}&\includegraphics[scale=0.4,trim=0 0 1cm 0.75cm,clip=true]{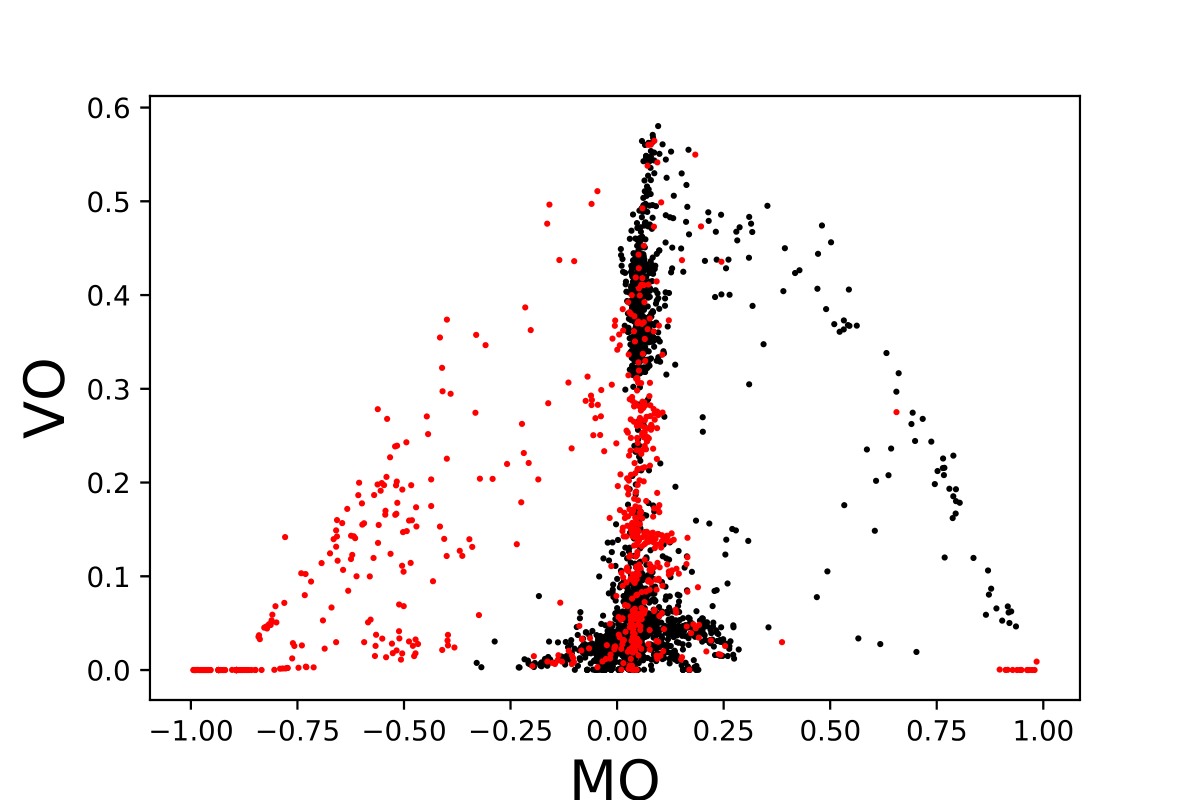}\\
 & \\
\hspace*{0.3cm}  FOM-fAO & \hspace*{0.3cm}  FOM-fSDO \\
\includegraphics[scale=0.4,trim=0 0 1cm 0.75cm,clip=true]{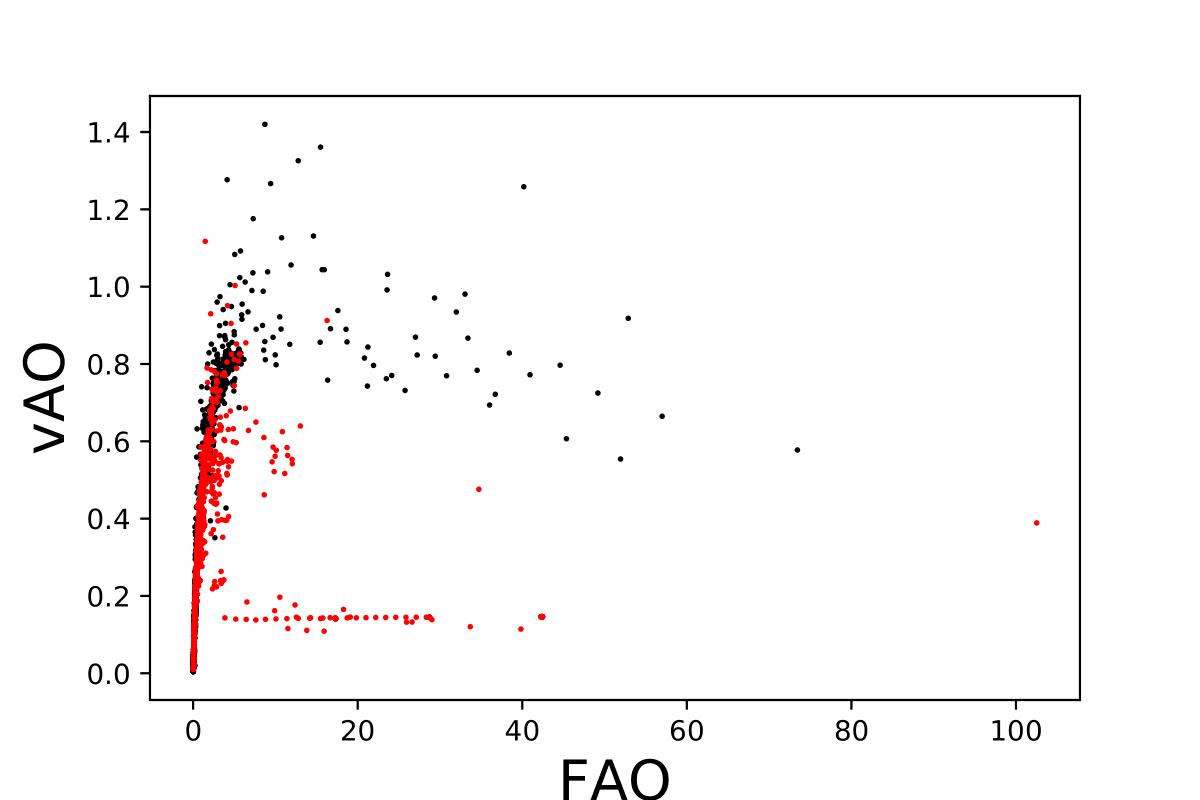}&\includegraphics[scale=0.4,trim=0 0 1cm 0.75cm,clip=true]{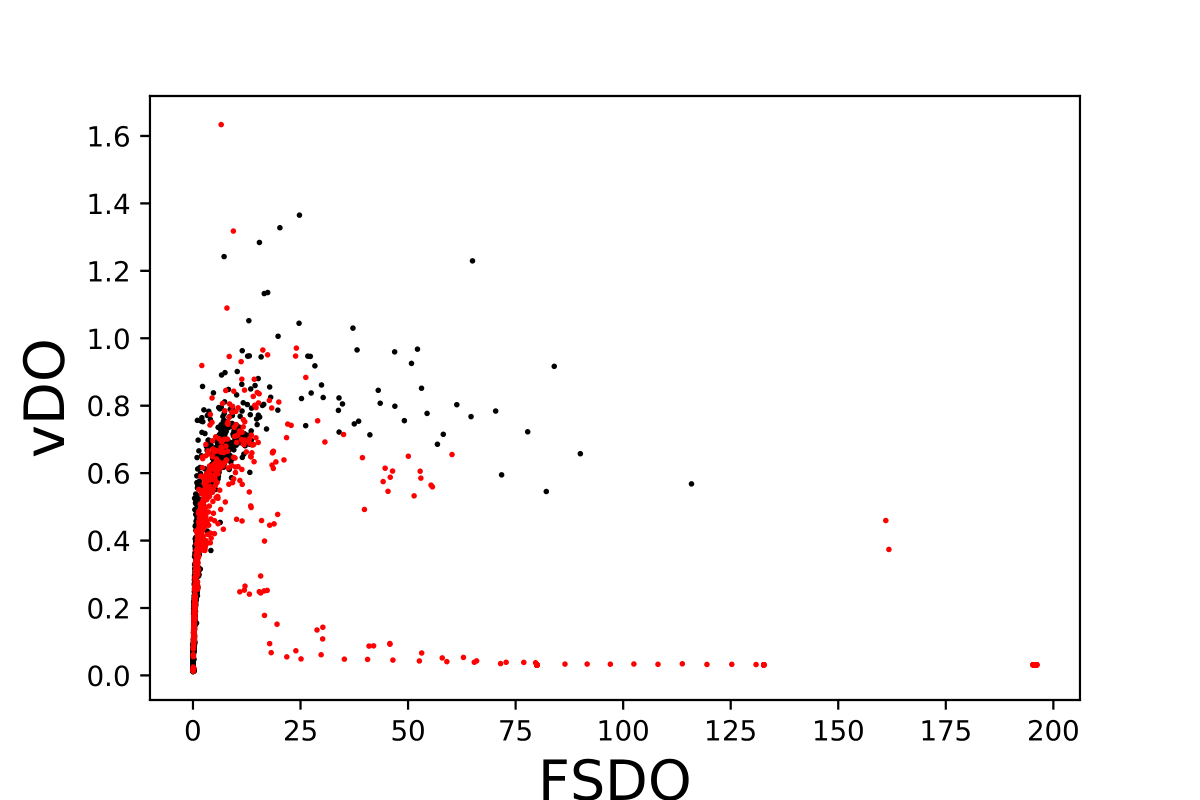} \\
\end{tabular}
\end{center}
\caption{Four visualization plots from Airbus data. In order from top to bottom and from left to right: FPCA, Magnitude-Shape plot with Integrated Tukey depth, Functional Outlier map with fAO and Functional Outlier map with fSDO. Black points corresponds to normal curves, red points to anomalies.}\label{visu}
\end{figure}

{\renewcommand{\arraystretch}{1.5}
{\setlength{\tabcolsep}{0.35cm}
\begin{table}[!h]
\vspace*{2cm}
\begin{center}
\begin{tabular}{c|cccc|||cccc}
\hline
& \multicolumn{4}{c}{\textbf{Airbus data}} &\multicolumn{4}{c}{\textbf{Rocks data}} \\ \hline
 Method & F1-Score &  AP & AUC& $p_c$ & F1-Score &  AP & AUC& $p_c$ \\ \hline
FIF &\textbf{0.81} &\textbf{0.88}& \textbf{0.76}&0.52 &0.974& \textbf{0.992}& 0.772& 0.10\\
fAO & 0.78&0.77&0.63&0.45&\textbf{0.989} & 0.991& \textbf{0.833}& \textbf{0.58}\\
fbd &0.78 &0.77&0.63&0.45&0.977& 0.988& 0.751& 0.19\\
fSDO &0.78 &0.77&0.63&0.45&0.972& 0.980& 0.555&0\\
fT & 0.78&0.77&0.63& 0.45& 0.984& 0.989& 0.780&0.43 \\
ACH &0.77 &0.77&0.62&0.44 & 0.972 & 0.961& 0.280&0 \\
Outliergram&0.71 &0.76 &0.55&0.28& 0.974 &0.981&0.66& 0.03  \\
MS + IF & 0.80& 0.76&0.64&0.51&  0.972 &0.981& 0.601&0  \\
FOM (fSDO) + IF & \textbf{0.81}&0.76 &0.66&0.53&0.972& 0.978& 0.530&0.02  \\
FOM (fAO) + IF &0.80& 0.77&0.65&0.51&  0.984& 0.991& 0.804&0.448  \\
FPCA + IF & \textbf{0.81} & 0.80& 0.70&0.53 &0.972&0.971& 0.446 & 0 \\
FPCA + LOF &0.72& 0.73&0.52&0.31&0.972&0.969& 0.445& 0.02\\
FPCA + OC&\textbf{0.81} &0.79&0.70&\textbf{0.54}&0.972&0.971& 0.463&0\\ \hline
\end{tabular}
\end{center}
\vspace*{0.2cm}
\caption{Methods considered in performance comparison with F1-Score, Average Precision (AP), AUC and sensitivity $(p_c)$ for Airbus  and Rocks data. Bold numbers correspond to the best result.}
\label{tab:bench}
\end{table}}}
\twocolumn

We analyze the complete Airbus dataset (\textit{i.e.} including normal and abnormal accelerometer data) as well as data arising from the spectrophotometry of rocks with wavelengths of light source ranging from $382$  to $930$ nanometers corresponding to visible-IR spectrum. 

The complete Airbus dataset is split into an unlabeled training dataset  ($1677$ sampled curves observed at $61440$ equally spaced time points)  used for the learning purposes and a labeled test dataset ($2511$ observations among which $717$ are labeled as abnormal) to evaluate performance using the metrics described in Section~\ref{metrics}. The second dataset contains linearly interpolated spectroscopy measurements of construction materials (mined rocks) from different locations and mining sites. Precisely, it includes $2038$ limestones and $58$ cell limes  of $600$ wavelength measures. 
Since the measured spectra are very noisy, and with the task of the current work being rather comparative (than absolute) performance of the methods, we modified the original dataset by removing most difficult to detect anomalies in a supervised manner (which only underlines the difficulty of the real-data anomaly detection task). More precisely, we removed $860$ normal data that have (on an average) same values as the average value of anomalies and that can not be distinguished by any of the algorithms. Such removal shall not prioritize any of the methods. We further perform a linear interpolation in order to obtain data with equidistant measurement points.


\subsection{Visualization}

While a simple plot of the rocks data provides relevant information on the data shape (see the Supplementary Material), the richness of the aeronautics curves makes it impossible. To get a first insight into the structure of the aeronautics dataset under study, we start with meaningful visualization. Recently, many graphical tools have been proposed in the literature, such as general-purpose functional highest density region plots~\cite{hyndman10}, functional boxplots~\cite{genton11} or amplitude and phase boxplot displays~\cite{geomvisua}, as well as those especially designed for anomaly detection such as the Outliergram~\cite{2014shape}, the Functional Outlier Map (FOM) \cite{hubert2015multivariate,FOM} or the Magnitude-Shape plot (MS)~\cite{MS}. This last group, together with the generic Functional Principal Component Analysis (FPCA) constitute our interest.


The Magnitude-Shape plot is two-dimensional outlyingness (or data depth)-based graphical tool, based on the outlyingness mean and variance---over the time domain---of the functional observation. For a sample of curves $\mathcal{C}_n=\{\bmX_1,...,\bmX_n\}$ observed on $\{t_1,\ldots, t_p\}$, the MS plot is then the scatter plot of the points $(\text{MO}_i,\text{VO}_i)_{1\leq i\leq n}$ with the Mean Outlyingness ($\text{MO}$)

$$\text{MO}_i=\frac{1}{p}\sum_{j=1}^p \; \text{O}(\bmX_i(t_j)\,\mid\,\mathcal{C}_n(t_j)),  $$

\noindent and the Variational Outlyingness ($\text{VO}$)

$$\text{VO}_i=\frac{1}{p}\sum_{j=1}^p \; \left( \text{O}(\bmX(t_j)\,\mid\,\mathcal{C}_n(t_j)) - \text{MO}_i  \right)^2,$$

\noindent where the outyingness itself is usually a monotone decreasing transform of data depth: $\text{O}(\bmX_i (t_j), \mathcal{C}_n(t_j))= \dfrac{1}{\text{D}(\bmX_i(t_j)\,\mid \,\mathcal{C}_n(t_j))-1}$.

\vspace*{0.3cm}
The Functional Outlier Map (FOM) is similar to the MS plot in the case of univariate functional data, with the only difference in measuring relative instead of absolute variability adapting thus to the actual variability of the function, see~\cite{FOM}. It is defined as a plot of points
$$\Bigl(\text{MO}_i, \frac{\sqrt{\text{VO}_i}}{1 + \text{MO}_i}\Bigr)_{1\leq i\leq n}.$$

For explainability purposes, vizualisation tools are computed on the test set where labels are available. Visualization plots, displayed in Figure~\ref{visu}, allow to identify certain anomalies (red points correspond to labeled anomalies), while others substantially overlap with the normal data (black points). They reveal the difficulty of marking the entity of abnormal observations for the auronautics data, and explain variations in the performance of different methods used in Section~\ref{bench}.

\subsection{Benchmark Study on Aeronautics and Rocks Data}\label{bench}

In Section~\ref{sim} only four types of anomalies were identified and studied in more detail, while many others remain that are not easy to associate with any existing taxonomy, which is often the challenge of real-world data. While many methods are designed for aiming at certain types of anomalies, clearly no universality in detecting abnormal observations of different types can be generally expected. 

To account for multiplicity of possible goals, we use several performance metrics: F1-Score, Average Precision (AP), AUC, and sensitivity $p_c$; see Section~\ref{metrics} for more details.

Table~\ref{tab:bench} displays the results. Methods perform differently and there is no general winner. First observation is the evidence of complexity of the real-world aeronautics dataset, which leads to non-perfect results across all the considered methods. This is also due to the fact that test data contains types of anomalies not present in the train data. Second, one should note that FIF appears to have very good performance, being best method in this benchmark in general. Even if its AUC of $76\%$ is not high, it can be still seen as satisfactory provided existence of $29\%$ of anomalies in the test data. Third, the depth-based methods indicate very stable results (mostly relatively satisfactory, except $p_c$), which comes from the fact that ordering of observations may be very similar for univarite functional curves due to (almost) coincidence of orderings for different univariate depths. While the rocks dataset was artificially simplified, the comparison remains similar and thus strengthens the conclusions.

Additionally, we display the ROC curves of the four best methods for both datasets (see Figure~\ref{roc3}).  The true positive rate is maximized by the FIF algorithm when the false positives are very close to zero which is crucial in many applications such as predictive maintenance in the case of airbus. This additional analysis makes FIF preferable to fAO in practice.


\begin{figure}[!h]\label{roc3}
\begin{center}
\begin{tabular}{c}
\includegraphics[scale=0.3,trim= 3cm 0 0 0]{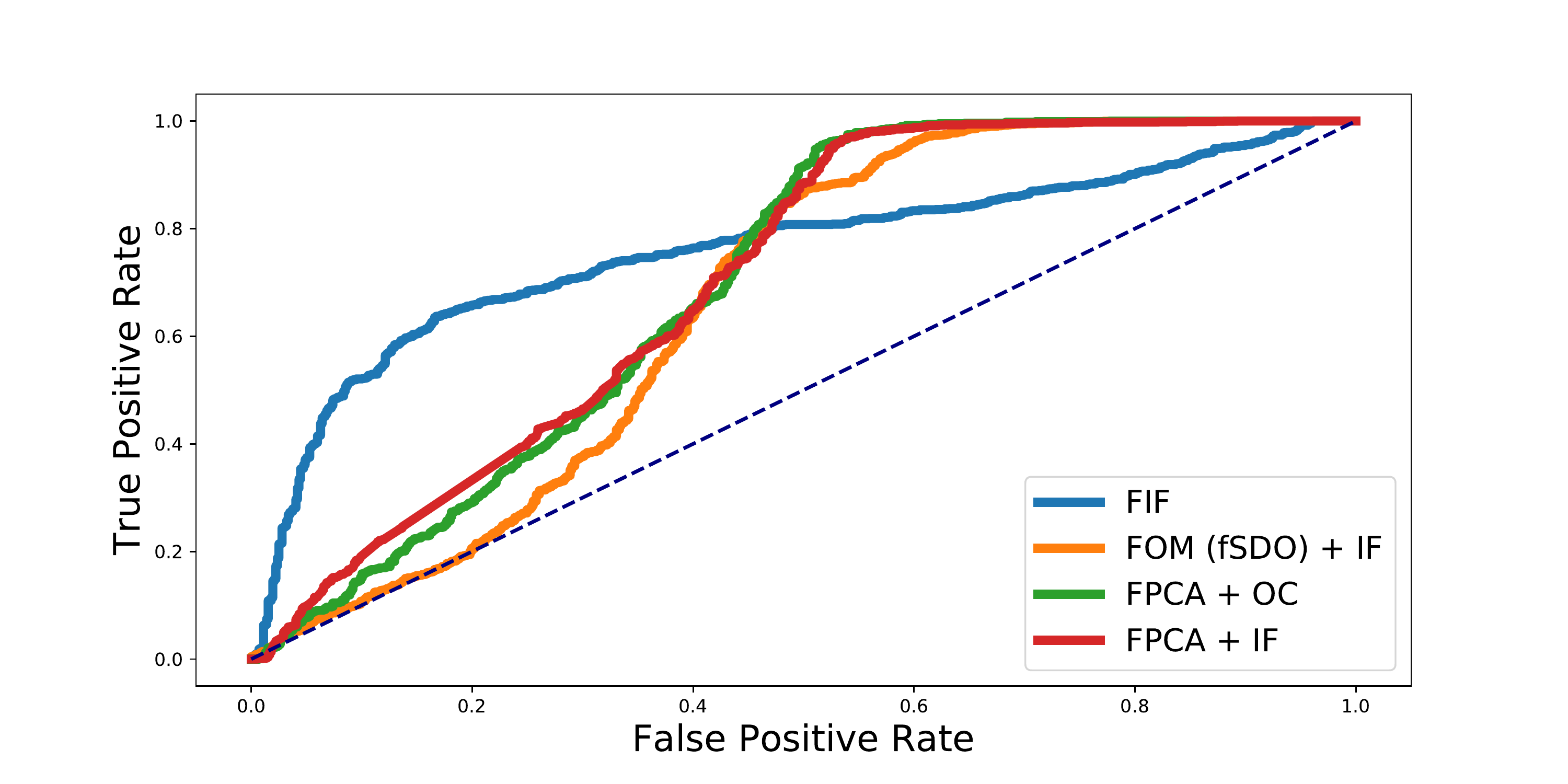}\\\includegraphics[scale=0.3,trim= 3cm 0 0 0 ]{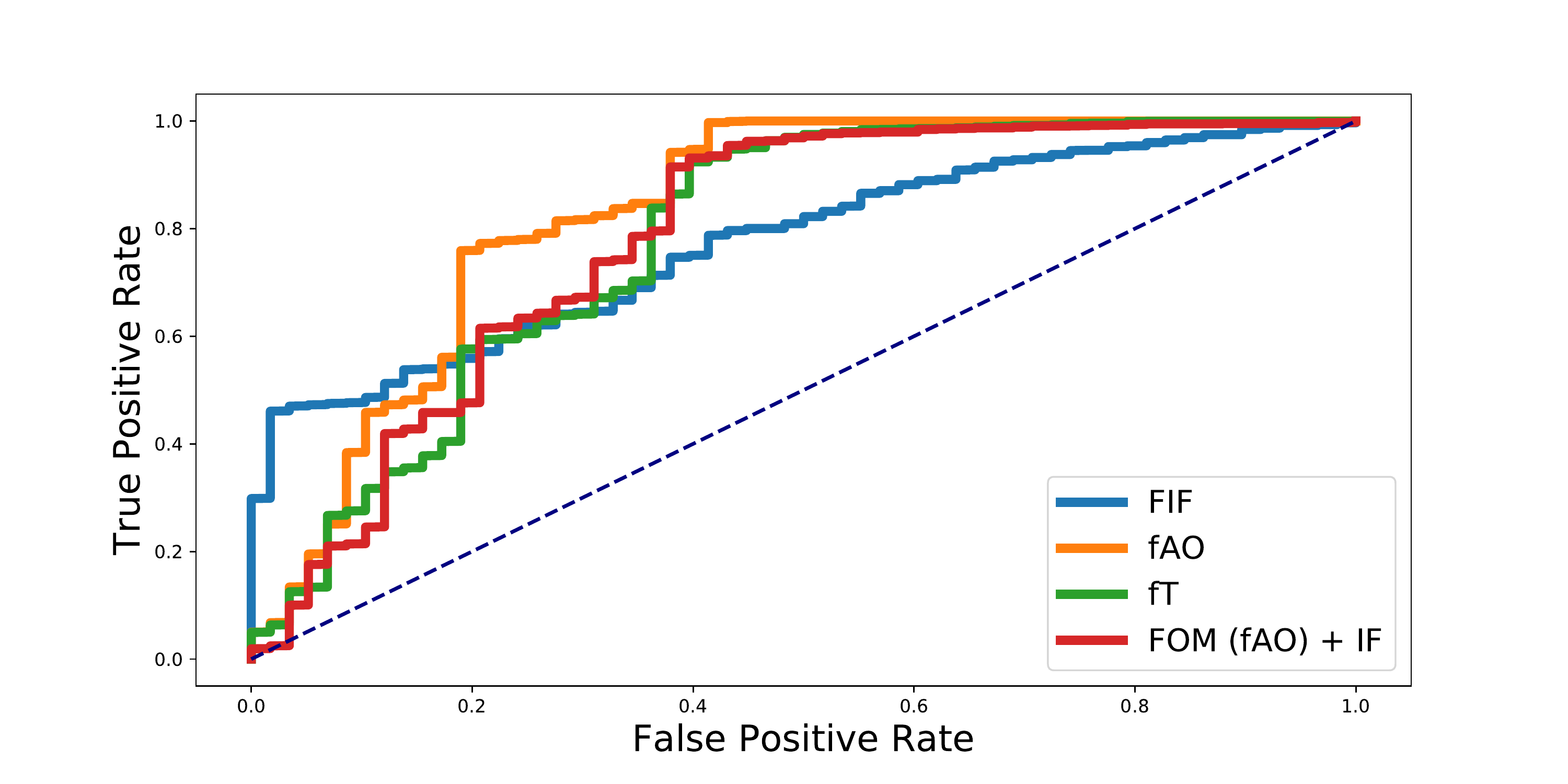}
\end{tabular}
\end{center}
\caption{ROC curves of the four best methods for the aeronautics (top) and rocks (bottom) data.}
\label{roc3}
\end{figure}

\section{Conclusion and Perspectives}\label{sec:concl}

Due to its unsupervised nature, the anomaly detection problem is extremely challenging. Because the learning algorithms used to build a decision rule in this context rely on 'normal' data solely, they mostly consist in identifying regions of the feature space where normal observations are less likely to fall in and cannot be based on empirical estimates of the performance metrics that shall be used afterwards to evaluate them when labels become available. This is even much more challenging in the case of observations valued in a vast functional space. In absence of prior knowledge about the types of the functional anomalies to be detected in the future, it is key to implement very flexible algorithms, exploiting the statistical information at disposal as far as possible. In this paper, we investigated the performance of recent anomaly detection techniques tailored to the functional framework such as Functional Isolation Forest and ACH depth, by means of real datasets, composed of sequences of accelerometer data measured on helicopters and spectrometry of construction materials. As confirmed by additional simulation studies, the benchmark revealed a clear advantage of these techniques compared to more traditional methods, relying on a preprocessing of the data (\textit{e.g.} FPCA). These very encouraging results call for further applications and extensions, to anomaly detection based on multivariate time-series in particular, the behavior of complex infrastructure being often continuously monitored by several sensors and not just one.

%

 \section*{Funding}
  This work has been funded by BPI France in the context of the PSPC Project Expresso (2017-2021). This project also received financial support from the initiative ``Forschungspartnerschaften Mineralrohstoffe - ein strategischer Forschungsschwerpunkt der Geologischen Bundesanstalt''. The spectroscopic data of sedimentary material was provided by the Geological Survey of Austria.

\bibliographystyle{unsrt}
\bibliography{fad_benchmark}

\clearpage
\begin{appendices}
\renewcommand\thefigure{\thesection.\arabic{figure}}

\section{Benchmarked datasets}

\begin{figure}[!h]
\begin{center}
\begin{tabular}{cc}
Aeronautics data & Rocks data\\
\includegraphics[scale=0.5]{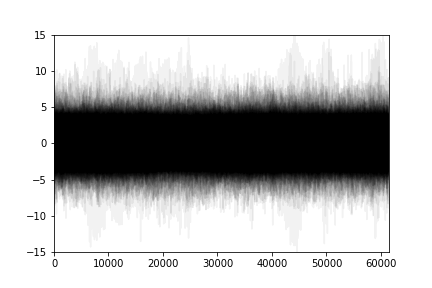} & \includegraphics[scale=0.5]{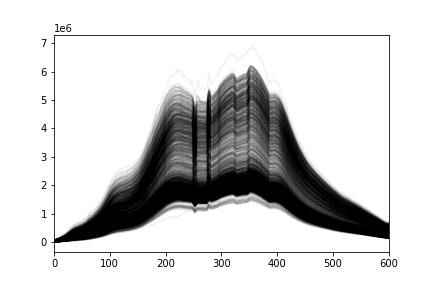}
\end{tabular}
\end{center}
\caption{The aeronautics and the rock datasets.}
\label{dataset}
\end{figure}

In this part, we display in Figure ~\ref{dataset} the aeronautics and the rocks datasets.

\section{Additional experiments on simulated anomalies}\label{additional}

In this part, complementary experiments to the Section~\ref{sec:simulation} are displayed. They are conducted  with the same methodology but varying proportion of anomalies: 1\% in Table ~\ref{tab:simu21}, 2\% in Table ~\ref{tab:simu22}, 3\% in Table ~\ref{tab:simu23} and  4\% in Table ~\ref{tab:simu24}.

\onecolumn
{\renewcommand{\arraystretch}{1.5} 
{\setlength{\tabcolsep}{0.5cm}
\begin{table}[!h]
\begin{center}
{\scriptsize
\begin{tabular}{p{2.5cm}cccccccc}
\hline
& \multicolumn{2}{c}{\textbf{Isolated} }&\multicolumn{2}{c}{\textbf{Magnitude I}}&\multicolumn{2}{c}{\textbf{Magnitude II} }&\multicolumn{2}{c}{\textbf{Shape}} \\
Methods & $p_c$ &AUC&$p_c$ &AUC&$p_c$ &AUC&$p_c$ &AUC\\ \hline
FIF &0 &0.20&\textbf{1}&\textbf{1}&0&0.32&\textbf{0.06}&\textbf{0.98}\\
fAO &0 &0.44&\textbf{1}&\textbf{1}&0&0.54&0&0.67\\
fbd &0 &0.44&\textbf{1}&\textbf{1}&0&0.54&0&0.68\\
fSDO & 0&0.42&\textbf{1}&\textbf{1}&0&0.54&0&0.77\\
fT& 0&0.43 &\textbf{1}&\textbf{1}&0&0.44&0& 0.72 \\
ACH&0 & \textbf{0.62}&\textbf{1}&\textbf{1}&\textbf{1}&\textbf{1}&0& 0.61 \\
Outliergram&0 &0.55&\textbf{1}&\textbf{1}&0&0.54&0& 0.51 \\
MS + IF &0& 0.05&\textbf{1} &\textbf{1}&0&0.77&0& 0.77 \\
FOM (fSDO) + IF &0 &0.27&\textbf{1}&\textbf{1}&\textbf{1}&\textbf{1}&0& 0.97 \\
FOM (fAO) + IF &0 &0.31&\textbf{1} &\textbf{1}&0.88&\textbf{1} &0&0.96 \\
FPCA + IF &0 &0.08&0&0.96&0& 0.77&0&0.86\\
FPCA + LOF &0&0.31& 0.18&0.86&0&0.42&0&0.67\\
FPCA + OC&0 &0.03&0&0.93&0&0.78&0&0.88\\ \hline
\end{tabular}}
\end{center}
\vspace*{0.2cm}
\caption{Methods considered in performance comparison with the sensitivity $(p_c)$ and the Area Under the Receiver Operating Characteristic (AUROC) for the four simulated models with 1\% of added anomalies.  Bold
numbers correspond to the best result. }
\label{tab:simu21}
\end{table}}}

{\renewcommand{\arraystretch}{1.5} 
{\setlength{\tabcolsep}{0.5cm}
\begin{table}[!h]
\begin{center}
{\scriptsize
\begin{tabular}{p{2.5cm}cccccccc}
\hline
& \multicolumn{2}{c}{\textbf{Isolated} }&\multicolumn{2}{c}{\textbf{Magnitude I}}&\multicolumn{2}{c}{\textbf{Magnitude II} }&\multicolumn{2}{c}{\textbf{Shape}} \\
Methods & $p_c$ &AUC&$p_c$ &AUC&$p_c$ &AUC&$p_c$ &AUC\\ \hline
FIF &0 &0.23&0.97&\textbf{1}&0&0.32&\textbf{0.26}&\textbf{0.98}\\
fAO &0 &0.44&\textbf{1}&\textbf{1}&0&0.54&0&0.67\\
fbd &0 &0.44&\textbf{1}&\textbf{1}&0&0.54&0&0.68\\
fSDO & 0&0.42&\textbf{1}&\textbf{1}&0&0.43&0&0.77\\
fT& 0&0.43 &\textbf{1}&\textbf{1}&0&0.44&0& 0.72 \\
ACH&0 & \textbf{0.62}&\textbf{1}&0.83&\textbf{0.94}&\textbf{1}&0& 0.61 \\
Outliergram&0 &0.55&\textbf{1}&\textbf{1}&0&0.54&0& 0.49 \\
MS + IF &0& 0.05&\textbf{1} &\textbf{1}&0&0.77&0& 0.77 \\
FOM (fSDO) + IF &0 &0.22&0.94&\textbf{1}&0.86&\textbf{1}&0& 0.95 \\
FOM (fAO) + IF &0 &0.26&0.89&\textbf{1}&0.66&0.99 &0&0.95 \\
FPCA + IF &0 &0.08&0&0.92&0& 0.71&0&0.90\\
FPCA + LOF &0&0.35& 0.09&0.81&0&0.48&0.09&0.77\\
FPCA + OC&0 &0.03&0&0.94&0&0.79&0&0.91\\ \hline
\end{tabular}}
\end{center}
\vspace*{0.2cm}
\caption{Methods considered in performance comparison with the sensitivity $(p_c)$ and the Area Under the Receiver Operating Characteristic (AUROC) for the four simulated models with 2\% of added anomalies.  Bold
numbers correspond to the best result. }
\label{tab:simu22}
\end{table}}}

{\renewcommand{\arraystretch}{1.5} 
{\setlength{\tabcolsep}{0.5cm}
\begin{table}[!h]
\begin{center}
{\scriptsize
\begin{tabular}{p{2.5cm}cccccccc}
\hline
& \multicolumn{2}{c}{\textbf{Isolated} }&\multicolumn{2}{c}{\textbf{Magnitude I}}&\multicolumn{2}{c}{\textbf{Magnitude II} }&\multicolumn{2}{c}{\textbf{Shape}} \\
Methods & $p_c$ &AUC&$p_c$ &AUC&$p_c$ &AUC&$p_c$ &AUC\\ \hline
FIF &0 &0.21&0.98&\textbf{1}&0&0.33&\textbf{0.49}&\textbf{0.98}\\
fAO &0 &0.44&\textbf{1}&\textbf{1}&0&0.54&0&0.67\\
fbd &0 &0.44&\textbf{1}&\textbf{1}&0&0.54&0&0.68\\
fSDO & 0&0.42&\textbf{1}&\textbf{1}&0&0.43&0&0.77\\
fT& 0&0.43 &\textbf{1}&\textbf{1}&0&0.44&0& 0.72 \\
ACH&0 & \textbf{0.60}&\textbf{1}&0.85&\textbf{0.88}&\textbf{1}&0& 0.60 \\
Outliergram&0 &0.55&\textbf{1}&\textbf{1}&0&0.54&0& 0.49 \\
MS + IF &0& 0.05&\textbf{1} &\textbf{1}&0&0.75&0& 0.77 \\
FOM (fSDO) + IF &0 &0.06&0.89&\textbf{1}&0.64&0.99&0& 0.89 \\
FOM (fAO) + IF &0 &0.24&0.81&\textbf{1}&0.70&0.99 &0&0.93 \\
FPCA + IF &0 &0.07&0&0.93&0& 0.70&0&0.93\\
FPCA + LOF &0&0.39& 0.16&0.87&0&0.57&0.17&0.70\\
FPCA + OC&0 &0.04&0&0.94&0&0.78&0&0.93\\ \hline
\end{tabular}}
\end{center}
\vspace*{0.2cm}
\caption{Methods considered in performance comparison with the sensitivity $(p_c)$ and the Area Under the Receiver Operating Characteristic (AUROC) for the four simulated models with 3\% of added anomalies.  Bold
numbers correspond to the best result. }
\label{tab:simu23}
\end{table}}}

{\renewcommand{\arraystretch}{1.5} 
{\setlength{\tabcolsep}{0.5cm}
\begin{table}[!h]
\begin{center}
{\scriptsize
\begin{tabular}{p{2.5cm}cccccccc}
\hline
& \multicolumn{2}{c}{\textbf{Isolated} }&\multicolumn{2}{c}{\textbf{Magnitude I}}&\multicolumn{2}{c}{\textbf{Magnitude II} }&\multicolumn{2}{c}{\textbf{Shape}} \\
Methods & $p_c$ &AUC&$p_c$ &AUC&$p_c$ &AUC&$p_c$ &AUC\\ \hline
FIF &0 &0.23&0.94&\textbf{1}&0&0.34&\textbf{0.58}&\textbf{0.98}\\
fAO &0 &0.44&\textbf{1}&\textbf{1}&0&0.54&0&0.67\\
fbd &0 &0.44&\textbf{1}&\textbf{1}&0&0.54&0&0.68\\
fSDO & 0&0.42&\textbf{1}&\textbf{1}&0&0.43&0&0.77\\
fT& 0&0.43 &\textbf{1}&\textbf{1}&0&0.44&0& 0.72 \\
ACH&0 & \textbf{0.63}&\textbf{1}&0.85&\textbf{0.85}&\textbf{1}&0& 0.60 \\
Outliergram&0 &0.55&\textbf{1}&\textbf{1}&0&0.54&0& 0.46 \\
MS + IF &0& 0.05&\textbf{1} &\textbf{1}&0&0.74&0& 0.77 \\
FOM (fSDO) + IF &0 &0&0.80&\textbf{1}&0.65&0.99&0& 0.87 \\
FOM (fAO) + IF &0 &0.10&0.89&\textbf{1}&0.55&0.98 &0&0.87 \\
FPCA + IF &0 &0.08&0&0.92&0& 0.72&0.28&0.96\\
FPCA + LOF &0&0.43& 0.26&0.82&0&0.59&0.18&0.70\\
FPCA + OC&0 &0.03&0&0.93&0&0.78&0.08&0.95\\ \hline
\end{tabular}}
\end{center}
\vspace*{0.2cm}
\caption{Methods considered in performance comparison with the sensitivity $(p_c)$ and the Area Under the Receiver Operating Characteristic (AUROC) for the four simulated models with 4\% of added anomalies.  Bold
numbers correspond to the best result. }
\label{tab:simu24}
\end{table}}}
\twocolumn

\end{appendices}



\end{document}